\pdfoutput=1

\documentclass[11pt]{article}

\usepackage[preprint]{acl}

\usepackage{times}
\usepackage{latexsym}
\usepackage{inconsolata}

\usepackage[T1]{fontenc}

\usepackage[utf8]{inputenc}
\usepackage[utf8]{inputenc} %
\usepackage[T1]{fontenc}    %
\usepackage{hyperref}       %
\usepackage{url}     
\usepackage{longtable}%
\usepackage{booktabs}       %
\usepackage{amsfonts}       %
\usepackage{nicefrac}       %
\usepackage{microtype}      %
\usepackage{xcolor}         %
\usepackage{enumitem}
\usepackage{multirow}

\usepackage{array}
\usepackage{booktabs}
\usepackage{xcolor}
\usepackage{colortbl}
\usepackage{graphicx}
\usepackage{amsmath}
\usepackage{CJKutf8}
\usepackage{caption}
\usepackage{subcaption}
\usepackage[normalem]{ulem}
\usepackage{float}
\usepackage{tikz}
\usetikzlibrary{topaths}
\tikzstyle{every picture}+=[remember picture,inner xsep=0,inner ysep=0.25ex]
\usepackage{tikz-dependency}
\usepackage{amsmath,stackengine}
\usepackage{pst-node}
\usepackage{soul}
\usepackage{setspace}

\definecolor{pastelpink}{RGB}{255,209,220}
\definecolor{pastelblue}{RGB}{173,216,230}
\definecolor{pastelgreen}{RGB}{152,251,152}
\definecolor{pastelyellow}{RGB}{255,255,153}
\definecolor{pastelpurple}{RGB}{221,160,221}
\definecolor{pastelorange}{RGB}{255,179,71}
\definecolor{pastelturquoise}{RGB}{175,238,238}
\definecolor{pastellavender}{RGB}{230,230,250}
\definecolor{pastelpeach}{RGB}{255,218,185}
\definecolor{pastelmint}{RGB}{208,240,192}
\definecolor{pastelred}{rgb}{1,0.8,0.8}

\newcommand{\hlred}[1]{\sethlcolor{pastelred}\hl{#1}}

\newcommand\hlcjk{\bgroup\markoverwith
  {\textcolor{pastelred}{\rule[-.5ex]{2pt}{2.5ex}}}\ULon}

\setstackgap{L}{9pt}
\newcounter{wordcnt}

\newcommand{\numwidth}{1.5em}

\newcommand{\wordcount}[2][0]{%
  \ifnum#1>-1\relax\setcounter{wordcnt}{#1}\fi%
  \wordcountaux#2 \relax%
  \setcounter{wordcnt}{0}} %

\def\wordcountaux#1 #2\relax{%
  \stepcounter{wordcnt}%
  \stackunder{#1}{\makebox[\numwidth][r]{\scriptsize\textcolor{red}{\thewordcnt}}}\ %
  \ifx\relax#2\relax\else\wordcountaux#2\relax\fi}

\newcommand{\rwordcount}[3][-1]{%
  \ifnum#1>-1\relax\setcounter{wordcnt}{#1}\fi%
  \setcounter{wordcnt}{#3} %
  \rwordcountaux#2 \relax%
  \setcounter{wordcnt}{-1}} %

\def\rwordcountaux#1 #2\relax{%
  \stackunder{#1}{\makebox[\numwidth][r]{\scriptsize\textcolor{red}{\thewordcnt}}}\ %
  \addtocounter{wordcnt}{-1}%
  \ifx\relax#2\relax\else\rwordcountaux#2\relax\fi}

\title{Disjoint Processing Mechanisms of Hierarchical and Linear\\Grammars in Large Language Models}

\author{Aruna Sankaranarayanan \\
  CSAIL, MIT\\
  \texttt{arunas@mit.edu} \\\And
  Dylan Hadfield-Menell \\
  CSAIL, MIT \\
  \texttt{dhm@mit.edu} \\\And
  Aaron Mueller \\
  Northeastern University\\
\texttt{aa.mueller@northeastern.edu}}

\begin{document}
\maketitle

\begin{abstract}
All natural languages are structured hierarchically. In humans, this structural restriction is neurologically coded: when two grammars are presented with identical vocabularies, brain areas responsible for language processing are only sensitive to hierarchical grammars. Using large language models (LLMs), we investigate whether such functionally distinct hierarchical processing regions can arise solely from exposure to large-scale language distributions. We generate inputs using English, Italian, Japanese, or nonce words, varying the underlying grammars to conform to either hierarchical or linear/positional rules. Using these grammars, we first observe that language models show distinct behaviors on hierarchical versus linearly structured inputs. Then, we find that the components responsible for processing hierarchical grammars are distinct from those that process linear grammars; we causally verify this in ablation experiments. Finally, we observe that hierarchy-selective components are also active on nonce grammars; this suggests that hierarchy sensitivity is not tied to meaning, nor in-distribution inputs. %
\end{abstract}

\section{Introduction}
In 1861, \citeauthor{broca-1861-area} found evidence that language processing functions are \emph{localized} in specific brain regions. Since then, our mapping of the brain has advanced tremendously; we now know that \textbf{functional specialization} can arise not only from biologically coded mechanisms, but also from experience \citep{baker-2007-specialization}. More recently, there has been significant interest in understanding the mechanisms of language processing in large language models \citep{olsson2022context,hanna2023how,yu-etal-2023-characterizing,todd2024function}, whose inductive biases are more general than those of humans.

Sensitivity and \textbf{functional selectivity} toward the hierarchical structure of language is a hallmark of human language processing \citep{chomsky1957syntactic,chomsky1965aspects}: brain regions selective towards hierarchical grammars are disjoint from regions selective towards linear structures, as well as hierarchical but non-linguistic structures such as those found in music or programming languages, or sentences constructed from nonce words \citep{malik2023constructed,federenko2016neural, ivanova2020comprehension,liu2020computer,varley2000evidence,varley2005agrammatic,apperly2006intact,fedorenko2016language,monti2009boundaries,fedorenko2011functional,amalric2019distinct,ivanova2021language,chen2023human}. Despite large quantities of evidence from humans for language selectivity, it is not clear whether language models would acquire similar selectivity from exposure to natural language data in the absence of human-like learning biases. \citet{kallini2024mission} recently find that autoregressive Transformer-based models \citep{vaswani2017attention} can more easily learn grammars that accord with the structures found in human language. While that study provides evidence from language acquisition, we are primarily interested in language processing in models trained on large text corpora.

Do large language models (LLMs) demonstrate distinct mechanisms for processing hierarchically structured vs.\ non-hierarchically structured sentences that are otherwise superficially identical? We derive inspiration from \citeposs{musso2003broca} experiment testing hierarchical and linear selectivity in human language processing. We replicate this experiment, to the extent possible,\footnote{\citeposs{musso2003broca} experiment required that subjects be fluent in their native language (German, in their case) but not have prior exposure to the foreign languages (Italian and Japanese) that they were tested on. We cannot fully satisfy this condition for LLMs, whose training distributions consist of  significant amounts of documents in non-English languages---though orders-of-magnitude fewer documents than for English.} on a series of large pretrained language models (\S\ref{sec:methods}). %
We design a series of superficially similar but structurally distinct grammaticality judgment tasks. We generate hierarchical grammars that accord to natural language structure, as well as linear grammars that are explained by positional insertion or transformation rules. 
Using these models and stimuli, we investigate the following research questions:

\textbf{RQ1.} Do models process hierarchically-structured inputs in a distinct manner from linearly-structured inputs? \emph{We find that LLMs demonstrate distinct behaviors and mechanisms for grammars defined by hierarchical versus linear structure.}

\textbf{RQ2.} Which model components are causally responsible for judging the grammaticality of hierarchical versus linear inputs? To what extent are these components shared? \emph{We find high overlap between hierarchical grammars, but low overlaps across hierarchical and linear grammars. Ablating hierarchy-sensitive components significantly reduces accuracy on hierarchical grammars, but affects accuracy on linear grammars to a significantly lesser extent.}

\textbf{RQ3.} Are findings from \textbf{RQ1} and \textbf{RQ2} dependent on grounding in the lexicon of the language(s) of the training corpus? Or do these distinctions also hold when given grammars generated using nonce words? \emph{We observe that the natural-language hierarchy-sensitive components also have significant influence on nonce grammars.}

These results provide evidence that model regions responsible for processing hierarchical linguistic structure are localizable and distinct. Further, these regions are selective for hierarchically structured language more broadly, and are not dependent on meaning nor in-distribution language inputs. This suggests that functional specialization toward hierarchical linguistic structure can arise solely from exposure to language data. Thus, even in the absence of strong human-like inductive biases, language-selective regions can emerge. %

\section{Methods}\label{sec:methods}

\subsection{Models}
We use Mistral-v0.3 (7B; \citealp{jiang2023mistral}), QWen 2 (0.5B and 1.5B; \citealp{yang2024qwen2technicalreport}), Llama 2 (7B; \citealp{touvron2023llama}), and Llama 3.1 (8B and 70B; \citealp{grattafiori2024llama3}). We select these models because they are open-weights, relatively commonly used, and are currently among the best-performing open models. In all experiments, we use nucleus sampling (temperature$=$ 0.1, $p=$ 0.9) to reduce variance. We run all experiments on a GPU cluster containing 4 A100s (80G each). We used approximately 1000 GPU hours during this study.

\subsection{Data}
\begin{table*}[!ht]
    \centering
    \resizebox{\textwidth}{!}{
        \begin{tabular}{lp{5.5cm}p{6cm}p{7.5cm}}
        \toprule
        & \textbf{Grammar} & \textbf{Positive Example} & \textbf{Negative Example}\\
        \midrule
        \parbox[t]{2mm}{\multirow{3}{*}{\rotatebox[origin=c]{90}{Hierarchical\ }}} & \textbf{Declarative}. Subject, verb, object. & a woman reads a chapter & a woman reads \tikz[baseline=(node1.base)]\node (node1){chapter}; \tikz[baseline=(node2.base)]\node (node2){a};
        \begin{tikzpicture}[overlay, remember picture]
             \draw[-latex,draw=red] (node2.north) to[bend right] (node1.north);
             \draw[-latex,draw=red] (node1.south) to[bend right] (node2.south);
        \end{tikzpicture} \\
        & \textbf{Subordinate}. Subject, verb taking a relative clause complement. & Sheela thinks that the woman reads the chapter & Sheela thinks that the woman reads \tikz[baseline=(node1.base)]\node (node1){chapter}; \tikz[baseline=(node2.base)]\node (node2){the};
        \begin{tikzpicture}[overlay, remember picture]
             \draw[-latex,draw=red] (node2.north) to[bend right] (node1.north);
             \draw[-latex,draw=red] (node1.south) to[bend right] (node2.south);
        \end{tikzpicture}\\
        & \textbf{Passive}. Like \textbf{Declarative}, but in the passive voice. & a chapter is read by a woman & a chapter is read by \tikz[baseline=(node1.base)]\node (node1){woman}; \tikz[baseline=(node2.base)]\node (node2){a}; \begin{tikzpicture}[overlay, remember picture]
             \draw[-latex,draw=red] (node2.north) to[bend right] (node1.north);
             \draw[-latex,draw=red] (node1.south) to[bend right] (node2.south);
        \end{tikzpicture}\\\midrule
        \multirow{5}{*}{\rotatebox[origin=c]{90}{Linear\ \ \ }} & \textbf{Negation}. Insert ``doesn't'' or ``don't'' at position 5. & \wordcount{a woman reads a \hlred{doesn't} chapter} &
        \wordcount{a woman reads a chapter \hlred{doesn't}}\\
        & \textbf{Inversion}. Invert the word order of \textbf{Declarative}. & \rwordcount{chapter a reads woman a}{5} & \rwordcount{chapter a reads}{5} \wordcount{\tikz[baseline=(node1.base)]\node (node1){a}; \tikz[baseline=(node2.base)]\node (node2){woman};} \begin{tikzpicture}[overlay, remember picture]
             \draw[-latex,draw=red] (node2.north) to[bend right] (node1.north);
             \draw[-latex,draw=red] (node1.south) to[bend right] (node2.south);
        \end{tikzpicture}\\
        & \textbf{Wh-word}. Insert wh-word at position 5. & \wordcount{did a woman reads a \hlred{when} chapter?} & \wordcount{did a woman reads a chapter \hlred{when}?}\\
        \bottomrule
        \end{tabular}
    }
    \caption{\textbf{Dataset.} List of grammars, descriptions of the rules defining each grammar, and positive (grammatical) and negative (ungrammatical) examples for each. We provide only English examples here for space; see App.~\ref{app:rule-desc} for descriptions and examples for all grammars.
    }
    \label{tab:en-template-examples}
\end{table*}

We define 3 classes of hierarchical and linear grammars respectively in English, Italian, and Japanese, yielding 18 grammars total. Each sentence is generated using templates inspired by the constructs defined in \citet{musso2003broca}. For each structure, we generate \textbf{positive} and \textbf{negative} examples. Each grammar, its underlying rule, and examples of corresponding positive and negative examples are available in Tables~\ref{tab:en-template-examples} and \ref{tab:all-template-examples}. Our dataset consists of 7 verbs with at least 5 subjects and objects each; fully enumerated, we have 1106 positive-negative example pairs for each grammar. We use a 50/50 train/test split (for 553 pairs for each grammar).\footnote{Data and code to replicate our experiments can be found on \href{https://github.com/arunasank/disjoint-processing-llms}{github}} 

The difference between hierarchical and linear grammars lies in whether their latent structure is explained via hierarchical or positional rules. Hierarchical grammars contain rules that conform to the hierarchical structure of natural language \citep{chomsky1957syntactic,everaert2015structures}. Linear grammars contain rules that are defined by word positions or relative word orderings---e.g., insert a word at position 4.
Such rules are argued to be impossible in human language \citep{chomsky1957syntactic,chomsky1965aspects}.

For each grammar, a \textbf{positive} example is one that conforms to the rule, and a \textbf{negative} example is one that does not. For all hierarchical grammars, we form the negative example by swapping the final two words of a positive example. For all linear grammars that entail inserting a word at a specific position, we form the negative example by inserting the word at the final position. For linear grammars that entail reversing the word order, we form the negative example by swapping the final two words after reversing the input. For \textbf{Italian last-noun agreement}, the positive example is created by agreeing the gender of the determiner of the first noun with the gender of the \emph{last} noun; we form negative examples by using a gender determiner that agrees with the first noun.

\section{Experiments}\label{sec:exps}

We run 4 experiments to evaluate the behaviors and mechanisms of 6 LLMs in processing hierarchical and linear grammars. We use an in-context learning setup. All models are pre-trained on datasets primarily consisting of English sentences, but also containing significant amounts of other high-resource languages. The exact training composition of most of these models is unknown; given that LLMs are typically trained on data extracted from the open web,\footnote{\href{https://web.archive.org/web/20231105192356/https://huggingface.co/mistralai/Mistral-7B-v0.1/discussions/8}{Relevant discussion on Huggingface}.} one can conjecture that the data would be composed largely of English text \citep{w3techs2024}, with large amounts of text from other common languages.
    
Experiment 1 compares the performance of all the pre-trained LLMs on grammaticality judgment tasks given hierarchical and linear grammars (\S\ref{sec:exp1}). Experiment 2 locates model components that are important for processing hierarchical and linear structures by treating hierarchical and linear inputs as counterfactuals (\S\ref{sec:exp2}). Experiment 3 investigates the causal role of these components by ablating them and then measuring changes in grammaticality judgment performance (\S\ref{sec:exp3}). Experiment 4 investigates whether the components identified in Experiment 2 merely distinguish grammars that are in-distribution with the training data, or show a more abstract universal sensitivity to hierarchical and linear structure on nonce sentences (\S\ref{sec:expt-4-jabberwocky}).

The input prompt in our in-context learning setup comprises ten demonstrations, followed by a test example (details in \S\ref{sec:exp1}). In all experiments, we conduct four trials, presenting mean results across four random seeds; demonstrations in the input prompt are randomized between trials, while test examples remain consistent. The format of the prompts remain consistent across experiments.
\begin{figure*}[!htp]
    \centering    \includegraphics[width=\textwidth]{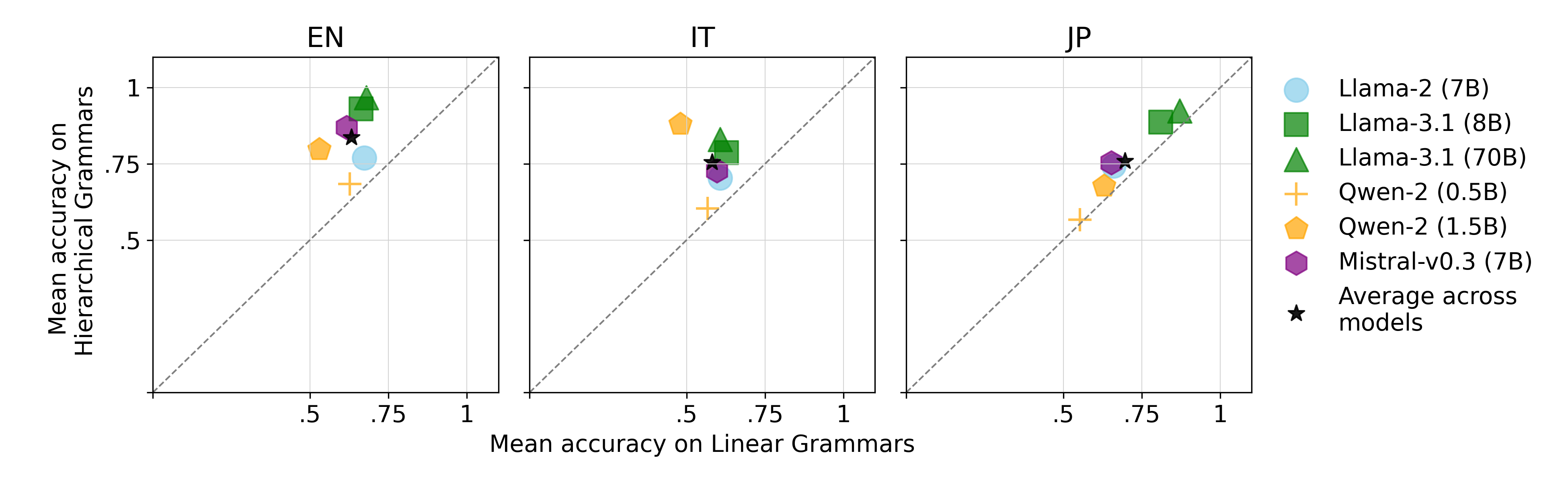}    
    \caption{Few-shot accuracy on the grammaticality judgment task on hierarchical and linear inputs. On average, all models are better at the grammaticality judgment task on hierarchical inputs as compared to linear inputs. On hierarchical grammars, models are best at processing English inputs followed by Italian and Japanese. Model-wise accuracy on this task is shown in Figure~\ref{fig:expt1-model-wise-bars} in App.~\ref{appendix:expt-1}. Grammar-wise accuracy is shown in Table~\ref{tab:expt1-model-accuracies-conv} in App.~\ref{appendix:expt-1}.
    }
    \label{fig:expt-1-few-shot-accuracy}
\end{figure*}

\subsection{Experiment 1: Are language models significantly more accurate at classifying the grammaticality of sentences from hierarchical grammars?}\label{sec:exp1}
We first evaluate the accuracy of LLMs on grammaticality judgments given examples from each grammar. \citet{musso2003broca} found that humans were more accurate at classifying examples of hierarchical grammars, even when they had no prior fluency in the test languages; we therefore hypothesize that a similar phenomenon would arise in LLMs if they contain functionally specialized regions for processing hierarchical structure.

As described in \S\ref{sec:methods}, for each of the 18 grammars, we generate 1106 examples. We perform a uniform 50/50 split to obtain our train/test split. Given a grammar, we first prompt an LLM with an instruction describing the nature of the in-context task (see Appendix~\ref{appendix:prompt}). This is followed by 10 demonstrations that are uniformly sampled from the training split. For each demonstration, we use the format ``\texttt{Q: \{sentence\}\textbackslash{}nA: \{answer\}}'', where \texttt{sentence} is the generated sentence, and \texttt{answer} is Yes if the sentence is a positive example, or No if it is a negative example. We ensure that each prompt contains exactly 5 positive and 5 negative examples; these can appear in any order. The model is then given the metalinguistic judgment task of generating ``\ Yes'' or ``\ No''  when given an example from the test split.\footnote{The inclusion of the leading space in the answer tokens is intentional, and conforms to the token that would have been included in the prompt if the answer were given to the model.} We extract the probabilities of the ``\ Yes'' and ``\ No'' tokens to determine if the model made the correct prediction. We report the accuracy of all the models in Figure~\ref{fig:expt-1-few-shot-accuracy}.

\paragraph{Hypothesis.}
Natural language is largely ambiguous with respect to linear versus hierarchical structure \citep{chomsky1957syntactic}; human brains have biological preferences for hierarchical structures \citep{musso2003broca}, but LLMs do not have this preference built into their architecture~\citep{min-etal-2020-syntactic,mccoy-2018-revisiting,mueller-etal-2022-coloring}, so it is not clear \emph{a priori} whether they would treat these structures in the same way. Given results from \citet{kallini2024mission}, we hypothesize that models will be significantly more accurate when labeling sentences from hierarchical than linear grammars. We also expect larger models to be more accurate.

\paragraph{Results.}
We find (Figure~\ref{fig:expt-1-few-shot-accuracy}) that for English and Italian grammars, models are better at distinguishing positive and negative examples in hierarchical grammars than linear grammars
($p$ < .001; see Table~\ref{tab:expt1-stat-sig}). This difference is greater for larger models than smaller ones, perhaps indicating greater functional specialization with scale. This provides initial support for our hypothesis that hierarchical and linear grammars are processed in distinct manners.

\subsection{Experiment 2: Are the model components implicated in processing hierarchical structures disjoint from those implicated in processing linear structures?}\label{sec:exp2}

Our behavioral evaluations suggest that LLMs are more accurate on grammaticality judgment tasks with hierarchical inputs, but this does not disambiguate whether models have separate \emph{mechanisms}\footnote{We use ``mechanism'' to refer to a causal chain proceeding from an initial cause to a final effect. In language models, this refers to a set of causally implicated model components that explain how inputs are transformed into the observed output behavior $m$, which we define below.} for processing hierarchical and linear grammars. If a model has specialized mechanisms for processing hierarchical and linear grammars, we hypothesize that the set of model components causally responsible for correct grammaticality predictions on hierarchical inputs should be different from those responsible for correct predictions on linear inputs.

To test this, we locate neurons in the model that are most sensitive towards processing hierarchical syntax. Specifically, we investigate dimensions of the output vector of the MLP and attention submodules in each layer.\footnote{For MLPs, we use the output of the down-projection \emph{after} the non-linear transformation. For attention, we use the output of the out projection.} We test whether there is significant overlap between the neurons responsible for processing hierarchical and linear structures.

Recall that we prompt the model with a task instruction followed by 10 uniformly sampled demonstrations of positive and negative examples. Given this prompt, we quantify the importance of each neuron $z$ in increasing the logit difference $m$ between the correct and incorrect answer tokens $y$ and $y'$ for a test sentence $t$. In other words, given a language model $\mathcal{M}$, $m = \mathcal{M}(t)_{y'} - \mathcal{M}(t)_y$; $\mathcal{M}(t)_y$ and $\mathcal{M}(t)_{y'}$ are the logits corresponding to the correct and incorrect answer tokens. We compute the component $z$'s \textbf{indirect effect} (IE; \citealp{pearl2001effects,robins1992indirect}) on $m$ given the test sentence $t$, and a minimally different sentence $t'$ that flips the correct answer from $y$ to $y'$.\footnote{If $t$ is a positive example, then $t'$ is the corresponding negative example formed by swapping the appropriate word(s) or modifying the sentence. If $t$ is a negative example, then $t'$ is the corresponding positive example.} Activation patching~\cite{vig2020causal,finlayson2021causal,geiger2020neural,meng2022locating}, a common procedure for computing the IE of model components, entails computing the IE as follows:  
\begin{equation}
    \text{IE}(m;z;t,t') = m(t | \text{do}(z_{t} = z_{t'}))  - m(t)
\end{equation}

Activation patching is computationally expensive, as the number of required forward passes scales linearly with the number of neurons. Therefore, we instead use attribution patching \cite{kramar2024atp, syed2023attribution}, a first-order Taylor approximation of the IE:
\begin{equation}
    \hat{\text{IE}}(m;z;t,t') = {\nabla_z m}|_{t} \left( z_{t'} - z_t \right)
\end{equation}
$\hat{\text{IE}}$ can be computed for \emph{all} $z$ using only 2 forward passes and 1 backward pass; i.e., the number of passes is constant with respect to the number of neurons. While not a perfect approximation, $\hat{\text{IE}}$ correlates almost perfectly with IE in typical cases \citep{kramar2024atp,marks2024sparse}.\footnote{Except at the first and last layer, where the correlation is still strong but significantly lower.}

We select the top 1\% of both attention and MLP neurons in the model by $\hat{\text{IE}}$. We compute the pairwise overlap of this top 1\% neuron subset for each pair of grammars to measure mechanistic overlap.

\paragraph{Hypothesis.} If there are distinct mechanisms for processing hierarchical and linear grammars, there should be significant overlap between pairs of hierarchical structures, and significant overlap between pairs of linear structures. However, overlaps \emph{across} linear and hierarchical structures should be significantly lower than overlaps between pairs of hierarchical grammars or pairs of linear grammars.

\begin{figure}[!ht]
    \centering
    \includegraphics[width=0.5\textwidth]{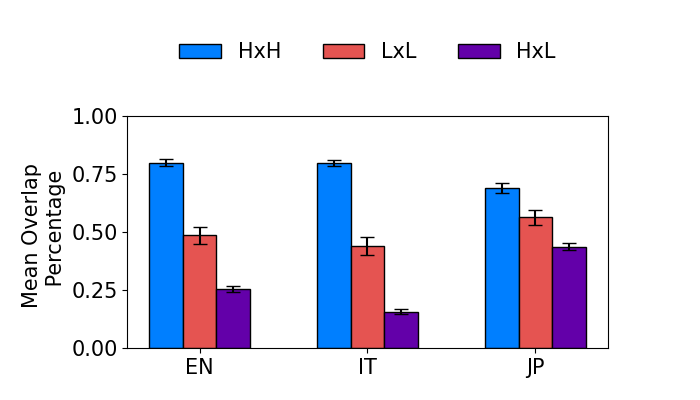}
    \caption{Mean pairwise overlap percentage of the top 1\% of neurons from hierarchical (H) or linear (L) grammars. We show means across models (error bars are standard errors); see Figure~\ref{fig:expt2-model-wise-overlaps-en-it-ja} in App.~\ref{appendix:exp2} for model-wise results. Overlaps are significantly (p $<$ 0.001, Table~\ref{tab:expt2-stat-sig-conv}) different between hierarchical-hierarchical pairs and linear-linear pairs, and between hierarchical-hierarchical pairs and hierarchical-linear pairs.}
    \label{fig:mean-overlap-on-conventional-tokens}
\end{figure}

\begin{figure}[!ht]
    \centering
    \includegraphics[width=0.5\textwidth]{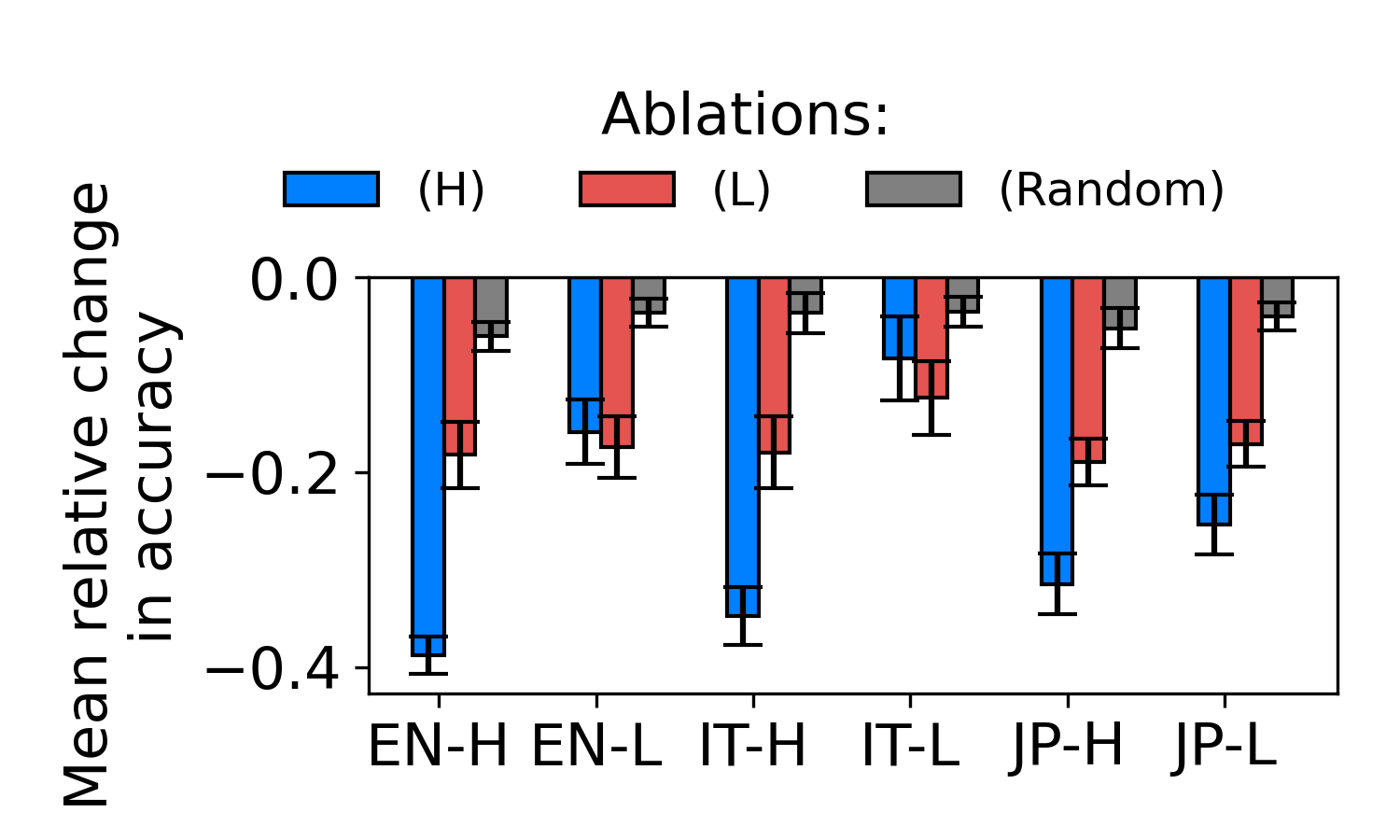}
    \caption{Mean relative change in accuracy across models (error bars are standard errors) after ablating the top 1\% of neurons from hierarchical (H) or linear (L) grammars. We compare to a random ablation baseline. For model-wise ablations, see Figure~\ref{fig:expt3-model-ablations-en-it-ja} in App.~\ref{appendix:exp3}.}
    \label{fig:mean-overlap}
\end{figure}

\paragraph{Results.}
We first observe that all mean pairwise component overlaps are significantly different from 0 (Figure~\ref{fig:mean-overlap}). However, this overlap is significantly higher ($p$ < 0.001) between pairs of hierarchical grammars than across pairs of hierarchical and linear grammars (See  Table~\ref{tab:expt2-stat-sig-conv} in App.~\ref{appendix:exp2} and Figure~\ref{fig:mean-overlap-on-conventional-tokens}). This holds across English, Italian, and Japanese. This supports the hypothesis that LLMs use specialized components for processing hierarchical syntax that are distinct from those responsible for processing linear syntax.

We also observe that linear structures that share a rule across languages, such as inversions, show stronger overlaps than arbitrary pairs of linear structures (Figures~\ref{fig:expt-2-conv-comp-mlp} and ~\ref{fig:expt-2-conv-comp-attn} in Appendix~\ref{appendix:exp2}). This serves as a sanity check that the component overlaps correlate with structural similarities.

\subsection{Experiment 3: Does ablating hierarchy-sensitive components affect performance on linear grammars, and vice versa?}\label{sec:exp3}
We have located neurons responsible for processing hierarchical and linear grammars. If these neurons are selective for only hierarchical or linear structure, then ablating them should selectively impact the model's performance on the grammaticality judgment tasks from \S\ref{sec:exp1}. We now perform an ablation experiment to causally verify this prediction.

Let $\bar{a_i}$ be the mean activation of neuron $a$ at token position $i$ across training examples. We first cache $\bar{a_i}$ for each MLP and attention output dimension. We then run three additional iterations of the grammaticality judgment task from \S\ref{sec:exp1}, each while ablating a different set of components. (i) We ablate the union of the top 1\% of neurons by $\hat{\text{IE}}$ across hierarchical grammars.
(ii) We take the union of the top 1\% of neurons across linear grammars, subsample to the same number of neurons as in the hierarchical union (subsampling procedure described below), and ablate this set. Finally, (iii) we ablate a random uniform subsample of neurons, where the number of ablated neurons is the same as in (i) and (ii). Sets (i) and (ii) are derived from \S\ref{sec:exp3}. We call the hierarchy-sensitive neuron set $H$ and the linearity-sensitive neuron set $L$. 

Due to the strong overlaps between components responsible for processing hierarchical syntax and only minimal overlaps between components responsible for processing linear syntax, we observe that $|L|\approx 2|H|$. We therefore subsample $L$ to be the same size as $H$ by (1) sorting components in $L$ by their effect size (as found in \S\ref{sec:exp3}) and (2) keeping the top $|H_\ell|$ components from layer $\ell$, where $|H_\ell|$ is the size of $H$ at layer $\ell$. When ablating the uniform subsample, we uniformly sample and ablate $|H_\ell|$ components in each layer $\ell$.

\paragraph{Hypothesis.} If $H$ and $L$ are functionally distinct, then ablating $H$ should reduce performance on hierarchical grammars more than ablating $L$ and more than ablating random components. Ablating $L$ should reduce performance on linear grammars more than ablating $H$ and more than ablating random components.

\paragraph{Results.}
Ablating components from $H$ decreases the models' accuracy on hierarchical structures significantly more than ablating components in $L$ (see Table~\ref{tab:exp3-stat-sig-conv} in App.~\ref{sec:exp3} for significance tests). Ablating components from $L$ decreases the model's accuracy on linear structures more than ablating $H$. Ablating uniformly sampled components causes a lower decrease in performance (if any) compared to ablations from the $H$ or $L$ sets.

These results are further mediated by model and language. The Llama-3.1 models as well as Mistral-v0.3 and QWen-2 (1.5B) show larger decreases in relative accuracy on hierarchical and linear inputs when ablating the $H$ and $L$ sets, respectively. Llama-2 and QWen-2 (0.5B) show similar changes in performance under ablations, though not in the selective manner we observe for other models. At the language level, these trends are consistent across English and Italian, but only sometimes generalize to Japanese. Overall, our results suggest that for Llama-3.1, Mistral-v0.3, and QWen (1.5B), the components discovered in \S\ref{sec:exp3} selectively reduce model performance in an expected manner in English and Italian. For other models, there is more mechanistic overlap in how grammatically judgments are performed for hierarchical and linear inputs. Thus, we largely find support for the hypothesis of hierarchical functional selectivity. Exceptions primarily include smaller models, and results in Japanese (a less frequent language in the training corpora of these models); this provides preliminary evidence that greater functional specialization may emerge with scale, both with respect to dataset size and number of parameters.

\subsection{Experiment 4: Are these neurons sensitive to hierarchical structure or in-distribution language?}\label{sec:expt-4-jabberwocky}

\begin{figure*}[ht]
    \centering
    \begin{subfigure}[b]{0.3\textwidth}
        \centering
        \includegraphics[width=\textwidth]{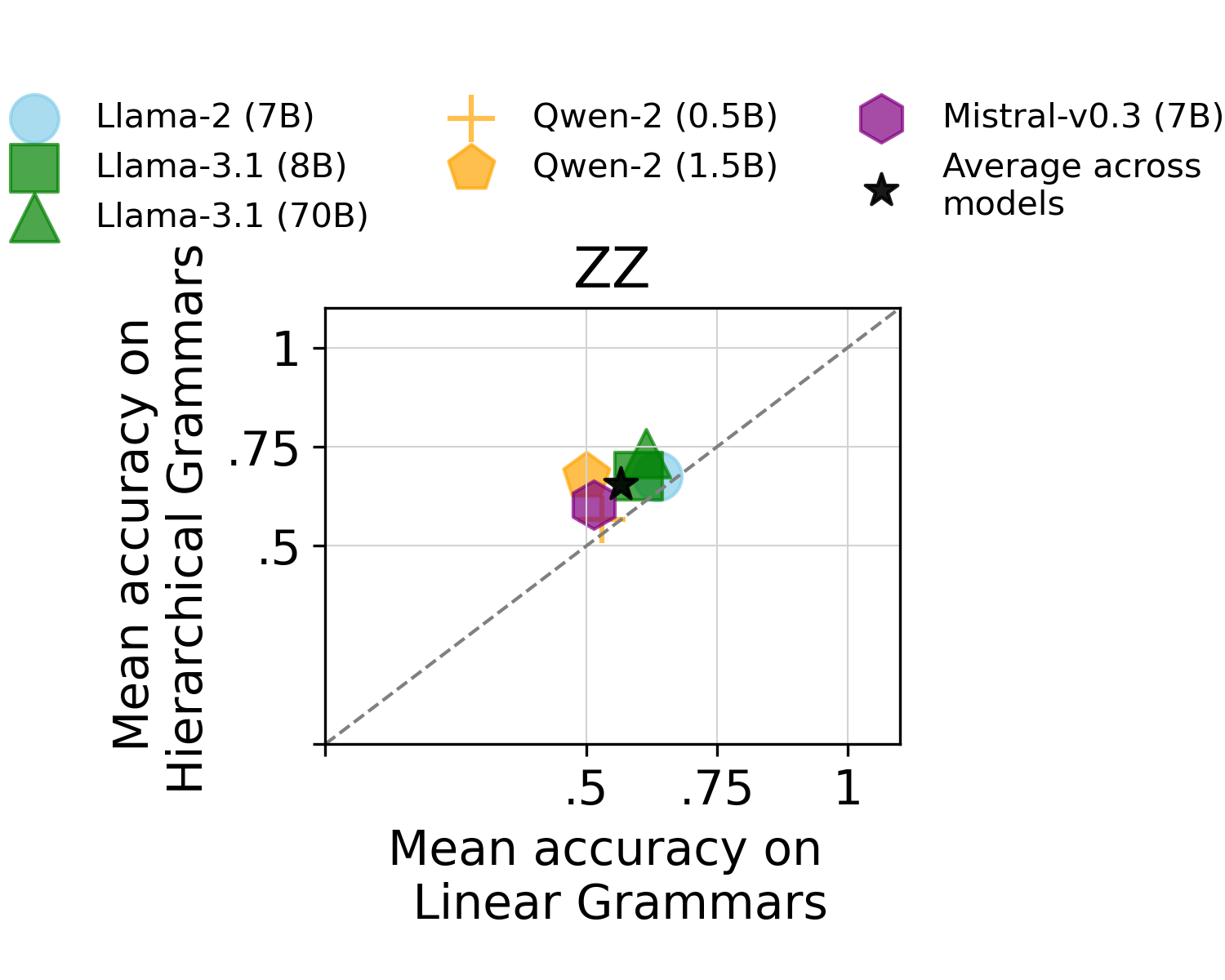}
        \caption{}
        \label{fig:expt1-jabber}
    \end{subfigure}
    \hspace{0.01\textwidth}
    \begin{subfigure}[b]{0.2\textwidth}
        \centering
        \includegraphics[width=\textwidth]{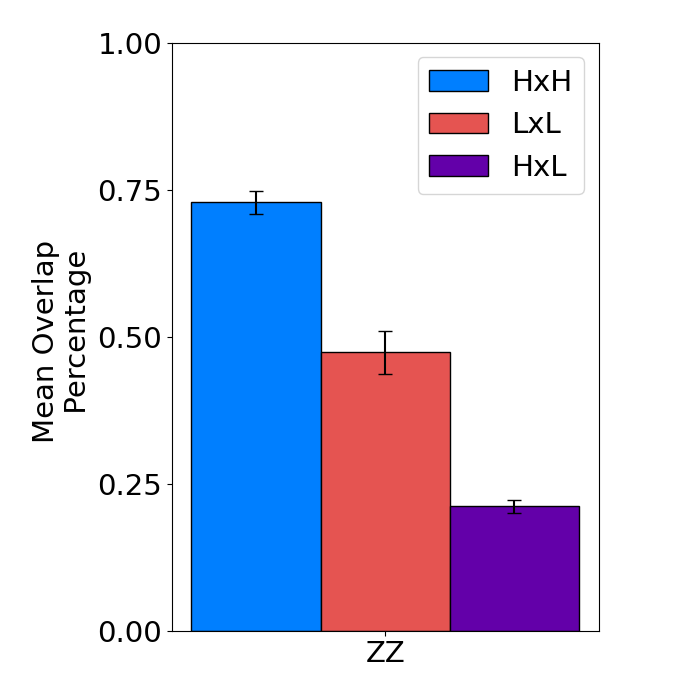}
        \caption{}
        \label{fig:expt2-jabber-within}
    \end{subfigure}
    \hspace{0.01\textwidth}
    \begin{subfigure}[b]{0.2\textwidth}
        \centering
        \includegraphics[width=\textwidth]{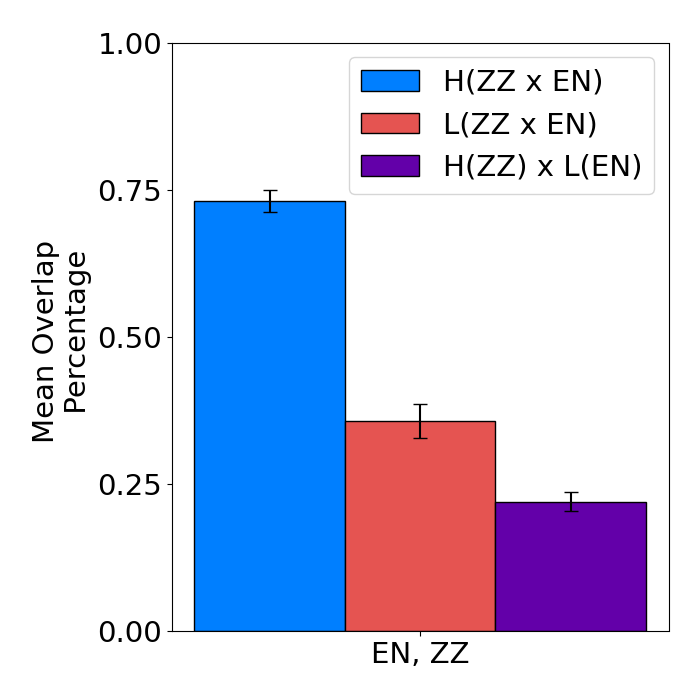}
        \caption{}
        \label{fig:expt2-en-jabber}
    \end{subfigure}
    \hspace{0.01\textwidth}
    \begin{subfigure}[b]{0.23\textwidth}
        \centering
        \includegraphics[width=\textwidth]{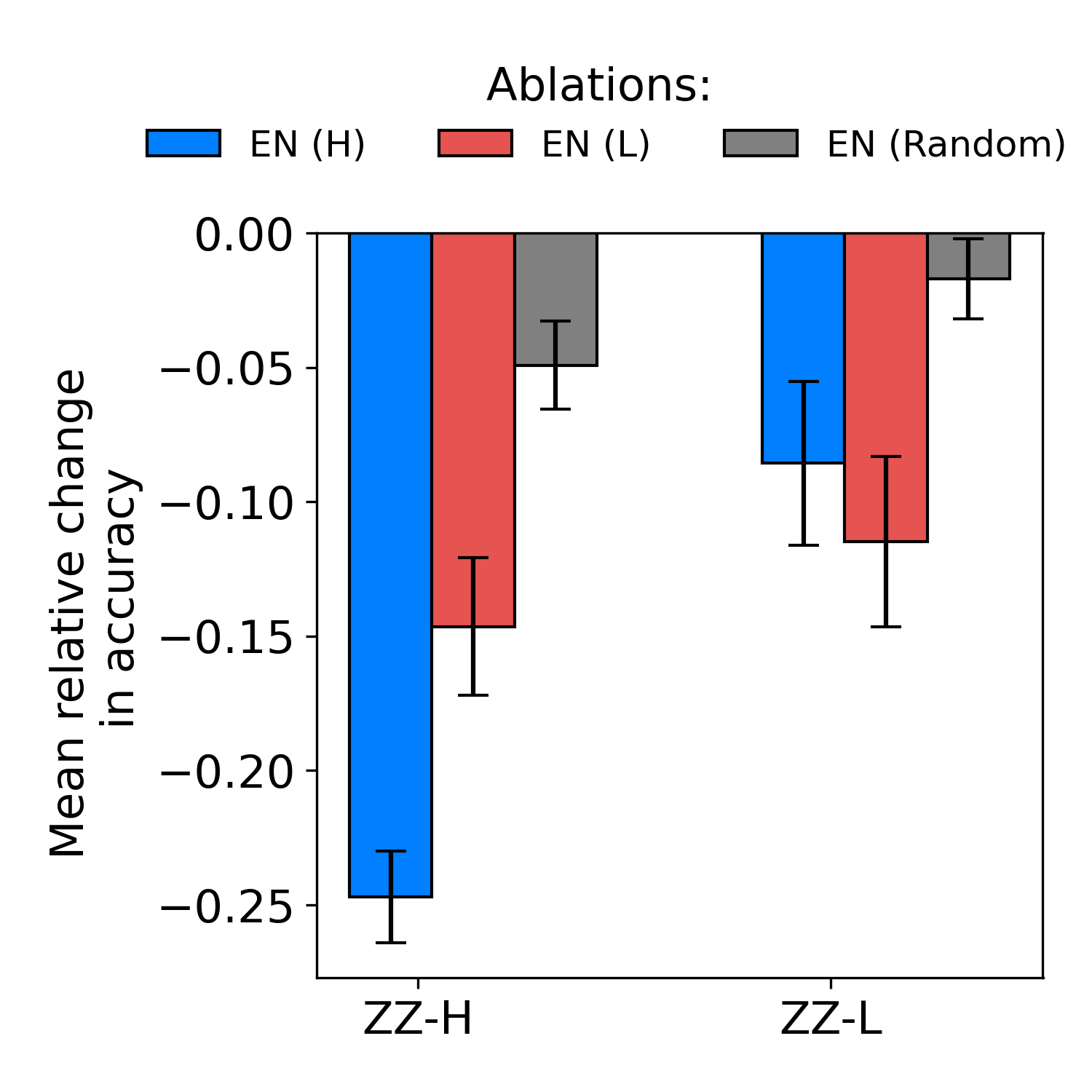}
        \caption{}
        \label{fig:jabber-conv-nonce-expt3}
    \end{subfigure}
    \caption{Results on Jabberwocky grammars. We show grammaticality judgment task performance (a), mean neuron overlap percentages between Jabberwocky hierarchical and linear grammars (b), neuron overlaps between English and Jabberwocky grammars (c), and the mean relative changes in accuracy as measured on Jabberwocky grammars after ablating top 1\% of neurons corresponding to English grammars (d). See App.~\ref{appendix:exp4} for model-wise results.}
    \label{fig:jabberwocky-figures}
\end{figure*}

Thus far, our results have been confounded by the fact that hierarchical sentences are commonly attested in the natural language corpora that LLMs are trained on, whereas linear sentences would be unlikely to appear. Thus, it is unclear if we have observed functional selectivity toward hierarchical language, or merely toward \emph{in-distribution} language. To address this confound, we propose additional experiments using sentences constructed from nonce words---what \citet{federenko2016neural} call \textbf{Jabberwocky sentences} (abbreviated ZZ).

We define a bijective mapping from all words in the English grammars to nonce words; see Table~\ref{tab:jab-template-examples} for examples. Then, we replicate our previous experiments on this set of out-of-distribution Jabberwocky sentences. By preserving the distinction between hierarchical and linear grammars and using a meaningless lexicon, we can disentangle hierarchy-sensitive mechanisms from mechanisms that are merely sensitive to natural language distributions resembling those in the training corpus.

\paragraph{Hypotheses.} In humans, Jabberwocky sentences cause a smaller increase in neural activity as compared to natural sentences \cite{federenko2016neural}, implying that the language processing regions of the brain are \emph{not} sensitive to Jabberwocky sentences.
If language models are similarly selective for meaningful inputs---and therefore, if the $H$ neurons from previous experiments are actually in-distribution-language neurons, and if the $L$ neurons are actually out-of-distribution language neurons---then we expect the following three trends. (1) There should not be significant differences in model performance on grammaticality judgments for hierarchical and linear Jabberwocky grammars. (2) There should be little overlap between the hierarchical English and Jabberwocky neurons; by implication, ablating neurons discovered from natural language inputs should not affect performance on Jabberwocky sentences. (3) It is not clear whether we should expect distinct mechanisms for processing H and L Jabberwocky grammars; if there is an abstract acceptability judgment circuit that is not tied to natural language, then it should be present in the linear natural-language neurons. Thus, we hypothesize that the $L$ neurons from previous experiments will affect performance on Jabberwocky sentences more than the $H$ neurons.

\paragraph{Results.} In behavioral experiments using Jabberwocky sentences, we find (Figure~\ref{fig:expt1-jabber}) that the gap in performance from hierarchical to linear grammars is significantly lower than that for English grammars---but still consistently present across models. The lower gap could be because performance is closer to chance than for natural grammars. The small gap in performance partially contradicts Hypothesis 1, but does not provide strong enough evidence to confidently reject.

Attribution patching results (Figure~\ref{fig:expt2-jabber-within}) suggest that the components used to correctly judge hierarchical and linear Jabberwocky inputs are largely disjoint: overlaps between pairs of hierarchical structures are significantly higher than overlaps across pairs of hierarchical and linear grammars. Moreover, the components used to process hierarchical English grammars are strongly shared with the components that are used to process hierarchical Jabberwocky grammars (Figure~\ref{fig:expt2-en-jabber}), while overlaps between linear English grammars and hierarchical Jabberwocky grammars is low. This contradicts Hypotheses 2 and 3, suggesting that the hierarchy-sensitive mechanisms we have observed in LLMs may be more abstract and generalized than those in humans.

Lastly, we observe (Figure~\ref{fig:jabber-conv-nonce-expt3}) that ablating the top 1\% of neurons from the English hierarchical grammars causes a significant decrease in accuracy when processing hierarchical Jabberwocky inputs; similar decreases in linear Jabberwocky accuracy result from ablating English linear components. This suggests that the causally relevant natural-language and Jabberwocky neuron sets are shared to a significant extent (see Table~\ref{tab:exp4-stat-sig-ablations} in App.~\ref{appendix:exp4}). Further, ablations to English hierarchical components causes \emph{selective} decreases in Jabberwocky hierarchical accuracy; selectivity is lower when ablating English linear components.

Taken together, these results provide evidence that LLMs' hierarchy-sensitive and linearity-sensitive component sets are sensitive primarily to the structure of the grammar, and only depend on grounding in meaning or in-distribution language to a minor extent. This provides significant ($p$ < .05) and causal evidence against Hypothesis 2, which we reject. Results from Figure~\ref{fig:jabber-conv-nonce-expt3} and Table~\ref{tab:exp4-stat-sig-ablations} also give sufficient evidence to reject Hypothesis 3. Thus, there exist grammaticality judgment mechanisms that are selective for hierarchical structure in a highly abstract manner, and that do not merely select for in-distribution language. That said, there are components are selective to both hierarchical \emph{and} in-distribution language, but these do not make up the majority of the components found in previous experiments.

\section{Discussion and Related Work}
\paragraph{Acquiring syntax-selective subnetworks.} We find behavioral and causal evidence supporting the hypothesis that hierarchical and linear grammars are processed using largely disjoint mechanisms in large language models. Thus, as in humans \citep{baker-2007-specialization}, general-purpose learners such as language models can \emph{acquire} functionally specific regions. To some extent, linguistic functional selectivity in LLMs is surprising: humans process many more modalities and signal types than language alone, so functional specialization toward linguistic signals may be sensible as one among many modal specializations \citep{kanwisher2010functional}. However, unimodal language models like those we test are exposed \emph{only} to text. While not all of this text is natural language, one might expect a larger portion of the model to be responsible for processing hierarchical structure. These as well as our behavioral results extend prior evidence that pretraining induces preferential reliance on syntactic features over positional features \citep{mueller-etal-2022-coloring,murty-etal-2023-grokking,ahuja2024learning},\footnote{Note, however, that these behavioral results may be explainable using teleological approaches such as those in \citet{mccoy2023embers}: linear grammaticality judgment is a low-probability task and contains low-probability inputs (assuming a pretraining distribution based on Internet text), and will therefore be more difficult for a language model to perform, even if the model used a shared mechanism to perform each grammaticality judgment task in this study.} and supports prior findings that there exist syntax-selective---and more broadly, language-selective---subnetworks in LLMs \citep{alkhamissi2024llmlanguagenetworkneuroscientific,sun2024brainlikefunctionalorganizationlarge}.

\paragraph{On human-likeness and learnability.} Note that hierarchical functional specialization is \emph{not} evidence that humans and LLMs process language in the same manner. \citet{federenko2016neural} find that language processing circuits in the brain activate significantly less on Jabberwocky sentences, whereas we observe significant overlaps (albeit not complete) in these circuits in LLMs. This suggests some degree of selectivity for natural in-distribution language, as in humans, but the hierarchy-sensitive mechanisms are also more abstract and not tied to meaning as in humans.

There is evidence that hierarchical grammars are easier to learn than grammars that do not occur in human languages \citep{kallini2024mission,ahuja2024learning}. This could provide an explanation for why language models are so attuned to this structure and learn to explicitly represent it: it is easier to learn a hierarchical organization than flat organization of a vocabulary, and it may simply be a more efficient explanation of the distribution. That said, randomly shuffling input data does not seem to destroy downstream performance \citep{sinha-etal-2021-masked}, despite destroying performance on structural probing tasks \citep{hewitt-manning-2019-structural}. Future work should investigate the relationship between the syntax-sensitive components we discover and performance on downstream NLP tasks.

\paragraph{Mechanistic interpretability.} Using causal localizations to investigate the mechanisms underlying model behaviors has recently become more popular \citep[e.g.,][]{wang2023interpretability,hanna2023how,prakash2024finetuning,merullo2024circuit,bayazit-etal-2024-discovering}. While localization is not equivalent to explanation, it can reveal distinctions in where and how certain phenomena are encoded in activation space. Future work could employ techniques from the training dynamics and mechanistic interpretability literature to better understand how and when these components arise during pretraining, as well as the (presumably numerous) functional sub-roles of these distinct component sets.

More broadly, this work suggests a less-explored direction in interpretability based on high-level coarse-grained abstractions. Much recent work has aimed to discover more fine-grained and single-purpose units of causal analysis (e.g., sparse autoencoder features; \citealp{bricken2023monosemanticity,cunningham2024sparse,marks2024sparse}); we believe that a parallel direction based in functionally coherent sets (or subgraphs) of components would yield equally interesting insights. For example, effective representations of syntax are a necessary condition for robust language understanding and generation; thus, we would expect the hierarchy-sensitive components we discover to be implicated in \emph{any} NLP task if the model were robustly understanding the inputs. Therefore, not relying on these components could be a signal that models have learned to rely on some mixture of heuristics.

\section{Conclusion}
We have investigated whether there exist localizable and functionally distinct sets of components for processing hierarchically versus linearly structured language inputs. We find behavioral and causal evidence that these component sets are distinct, both in location and in their functional role in the network.

\section*{Limitations}
We acknowledge that our work could be improved in several respects. First, neurons and attention outputs are problematic units of analysis due to polysemanticity \citep{elhage2022superposition}; i.e., observing the activations of a component is often not informative, as they are sensitive to many features simultaneously. Further, the component sets we analyze are unordered sets, which means that we do not yet understand how many distinct mechanisms are responsible for the behaviors we observe, nor what these mechanisms qualitatively represent. We have also not evaluated the effect of these components on tasks outside of grammaticality judgments; thus, we do not yet understand how selective nor how robust these behaviors or localizations are under different settings. 

Second, the grammaticality judgment task may prime the model to be sensitive to valid linguistic structures more generally, rather than the structures that we present to the models. Therefore, we cannot confidently conclude that the significant accuracy differences we observe will generalize to other task settings or prompt formats given the same grammars.

\bibliography{latex/custom}
\appendix
\newpage
\section{Methods}
\subsection{Grammar rule descriptions}
We define a series of hierarchical sentences in English, Japanese, and Italian. \label{app:rule-desc}
\begin{itemize}
    \item \textbf{Declarative sentence}:  For English sentences, subjects and objects can be singular or plural nouns. Verbs agree with their subjects. 
    $IT$ sentences are Italian translations of the English sentences. Unlike Italian and English which have SVO word order, Japanese translations ($JP$ sentences) have SOV word order. 
    \item \textbf{Subordinate sentence}: In each language, matrix subjects, subordinate subjects, matrix objects, and subordinate objects can be singular or plural nouns. In English and Italian,  verbs of the subordinate subject and the subject agree with their respective subjects in number. We generate subordinate clauses by using verbs which take complementizer phrases as objects (e.g., ``Tom sees that the dog carries the fish''). English and Italian both place the main clause's verb before the start of the subordinate clause, whereas Japanese places the main verb after the end of the clause.
    \item \textbf{Passive sentence}: Subjects and objects can be singular or plural nouns. Verbs are always in the passive form. Like in \cite{musso2003broca}, in the passive construction, we include the agent of a transitive verb in a prepositional phrase. 
    \item \textbf{Null subject sentence}: This structure is restricted to Italian. 
    We use the verb and object without the subject, since the use of the subject is not a strict requirement in Italian.\footnote{Italian verbal morphology provides all person and number information needed to understand the subject of a sentence, whereas English morphology does not provide this information. That said, there exist languages without the verbal person/number inflection that optionally allow dropping the subject of the sentence if it is the topic of that sentence, such as Mandarin and Japanese; thus, this structure is still attested and therefore still qualifies as a hierarchical (UG-compliant) structure.}
\end{itemize}

\paragraph{Linear Grammars} Similar to \citet{musso2003broca}, the linear sentences we test are constructed by breaking the hierarchical order between the subject and the nominal words. While our linear sentences use English, Italian, and Japanese lexicons, they break the hierarchical relationship between the subject, verb, and object, using the strategies described below.
\begin{itemize}
    \item \textbf{Negation}: We break the hierarchical order by inserting a negation word ``doesn\'t'' after the fifth word in English sentences. In Italian, we insert 
    `non' ($IT$) after the third word. In Japanese, we insert
    \begin{CJK}{UTF8}{min}ない\end{CJK} ($JP$) after the third word.
    \item \textbf{Inversion}: We invert the order of the words in a sentence (before tokenization) to form the second construction.
\end{itemize}

The third construction varies between languages. 
\begin{itemize}
    \item \textbf{Wh-word} (English): We include a question in the subordinate clause of the sentence by inserting a `wh-' word (who, why, what etc.) at the penultimate token position.
    \item \textbf{Last noun agreement} (Italian): We change the subject term's gender to always match that of the final noun in the sentence.
    \item \textbf{Past Tense} (Japanese): The Japanese past tense construction was built by adding the suffix -ta, not on the verb element as in the hierarchical grammatical rule for Japanese, but on the third word, counting from right to left.
\end{itemize}

\subsection{Dataset Description and Examples}\label{app:dataset-desc}
Examples of all the grammars we construct in English, Italian, and Japanese may be found in Table~\ref{tab:all-template-examples}. Examples of Jabberwocky sentences may be found in Table~\ref{tab:jab-template-examples}.
\begin{table*}[!h]
    \centering\renewcommand{\arraystretch}{2}
    \resizebox{\textwidth}{!}{
        \begin{tabular}{llp{5.5cm}p{6cm}p{7.5cm}}
        \toprule
        & \textbf{Language} & \textbf{Grammar} & \textbf{Positive Example} & \textbf{Negative Example}\\
        \midrule
        \parbox[t]{2mm}{\multirow{11}{*}{\rotatebox[origin=c]{90}{Hierarchical}}} & \multirow{3}{*}{English (EN)}  & \textbf{Declarative} & a woman reads a chapter & a woman reads \tikz[baseline=(node1.base)]\node (node1){chapter}; \tikz[baseline=(node2.base)]\node (node2){a};
        \begin{tikzpicture}[overlay, remember picture]
             \draw[-latex,draw=red] (node2.north) to[bend right] (node1.north);
             \draw[-latex,draw=red] (node1.south) to[bend right] (node2.south);
        \end{tikzpicture} \\
        & & \textbf{Subordinate} & Sheela thinks that the woman reads the chapter & Sheela thinks that the woman reads \tikz[baseline=(node1.base)]\node (node1){chapter}; \tikz[baseline=(node2.base)]\node (node2){the};
        \begin{tikzpicture}[overlay, remember picture]
             \draw[-latex,draw=red] (node2.north) to[bend right] (node1.north);
             \draw[-latex,draw=red] (node1.south) to[bend right] (node2.south);
        \end{tikzpicture}\\
        & & \textbf{Passive} & a chapter is read by a woman & a chapter is read by \tikz[baseline=(node1.base)]\node (node1){woman}; \tikz[baseline=(node2.base)]\node (node2){a}; \begin{tikzpicture}[overlay, remember picture]
             \draw[-latex,draw=red] (node2.north) to[bend right] (node1.north);
             \draw[-latex,draw=red] (node1.south) to[bend right] (node2.south);

             \draw[black!30, line width=0.5mm] (-19.5, -0.3) -- (2.6, -0.3);
        \end{tikzpicture}\\

        & \multirow{3}{*}{Italian (IT)}      
        & \textbf{Declarative} & una donna legge un capitolo & una donna legge \tikz[baseline=(node1.base)]\node (node1){capitolo}; \tikz[baseline=(node2.base)]\node (node2){un}; \begin{tikzpicture}[overlay, remember picture]
             \draw[-latex,draw=red] (node2.north) to[bend right] (node1.north);
             \draw[-latex,draw=red] (node1.south) to[bend right] (node2.south);
        \end{tikzpicture} \\
        & & \textbf{Subordinate} & Sheela pensa che una donna legge un capitolo &
        Sheela pensa che la donna legge \tikz[baseline=(node1.base)]\node (node1){capitolo}; \tikz[baseline=(node2.base)]\node (node2){un}; \begin{tikzpicture}[overlay, remember picture]
             \draw[-latex,draw=red] (node2.north) to[bend right] (node1.north);
             \draw[-latex,draw=red] (node1.south) to[bend right] (node2.south);
        \end{tikzpicture}\\
        & & \textbf{Passive} & un capitolo è letto da una donna & un capitolo è letto da \tikz[baseline=(node1.base)]\node (node1){donna}; \tikz[baseline=(node2.base)]\node (node2){una}; \begin{tikzpicture}[overlay, remember picture]
             \draw[-latex,draw=red] (node2.north) to[bend right] (node1.north);
             \draw[-latex,draw=red] (node1.south) to[bend right] (node2.south);
             \draw[black!30, line width=0.5mm] (-19.9, -0.3) -- (2.2, -0.3);
        \end{tikzpicture}\\
        & \multirow{3}{*}{Japanese (JP)} & \textbf{Declarative} 
        & \begin{CJK}{UTF8}{min}女性 は 章 を 読む\end{CJK} & 
        \begin{CJK}{UTF8}{min}女性 は 章\end{CJK} \tikz[baseline=(node1.base)]\node (node1){\begin{CJK}{UTF8}{min}読む\end{CJK}}; \tikz[baseline=(node2.base)]\node (node2){\begin{CJK}{UTF8}{min}を\end{CJK}}; \begin{tikzpicture}[overlay, remember picture]
             \draw[-latex,draw=red] (node2.north) to[bend right] (node1.north);
             \draw[-latex,draw=red] (node1.south) to[bend right] (node2.south);
        \end{tikzpicture} \\
        & & \textbf{Subordinate} & \begin{CJK}{UTF8}{min}シーラ は 女性 が 章 を 読む と 考える\end{CJK} & \begin{CJK}{UTF8}{min}シーラ は 女性 が 章 を 読む \end{CJK} \tikz[baseline=(node1.base)]\node (node1){\begin{CJK}{UTF8}{min}考える\end{CJK}}; \tikz[baseline=(node2.base)]\node (node2){\begin{CJK}{UTF8}{min}と\end{CJK}}; \begin{tikzpicture}[overlay, remember picture]
             \draw[-latex,draw=red] (node2.north) to[bend right] (node1.north);
             \draw[-latex,draw=red] (node1.south) to[bend right] (node2.south);
        \end{tikzpicture}\\
        & & \textbf{Passive} & \begin{CJK}{UTF8}{min}章 は 女性 に 読まれる\end{CJK} & \begin{CJK}{UTF8}{min}章 は 女性\end{CJK} \tikz[baseline=(node1.base)]\node (node1){\begin{CJK}{UTF8}{min}読まれる\end{CJK}}; \tikz[baseline=(node2.base)]\node (node2){\begin{CJK}{UTF8}{min}に\end{CJK}}; \begin{tikzpicture}[overlay, remember picture]
             \draw[-latex,draw=red] (node2.north) to[bend right] (node1.north);
             \draw[-latex,draw=red] (node1.south) to[bend right] (node2.south);
        \end{tikzpicture} \\
        \midrule\midrule%
        \multirow{14}{*}{\rotatebox[origin=c]{90}{Linear}} & \multirow{5}{*}{English (EN)} & \textbf{Negation}. Insert ``doesn't'' or ``don't'' at position 5. & \wordcount{a woman reads a \hlred{doesn't} chapter} &
        \wordcount{a woman reads a chapter \hlred{doesn't}}\\
        & & \textbf{Inversion}. Invert the declarative word order. & \rwordcount{chapter a reads woman a}{5} & \rwordcount{chapter a reads}{5} \wordcount{\tikz[baseline=(node1.base)]\node (node1){a}; \tikz[baseline=(node2.base)]\node (node2){woman};} \begin{tikzpicture}[overlay, remember picture]
             \draw[-latex,draw=red] (node2.north) to[bend right] (node1.north);
             \draw[-latex,draw=red] (node1.south) to[bend right] (node2.south);
        \end{tikzpicture}\\
        & & \textbf{Wh-word}. Insert wh-word at position 5. & \wordcount{did a woman reads a \hlred{when} chapter?} & \wordcount{did a woman reads a chapter \hlred{when}?} 
        \begin{tikzpicture}[overlay, remember picture]
             \draw[black!30, line width=0.5mm] (-21.5, -1.2) -- (0.2, -1.2);
        \end{tikzpicture}\\

        & \multirow{5}{*}{Italian (IT)} & \textbf{Negation}. Insert ``no'' at position 5. & \wordcount{una donna legge un \hlred{no} capitolo} & \wordcount{una donna legge un capitolo \hlred{no}}\\
        & & \textbf{Inversion}. Invert the declarative word order. & \rwordcount{capitolo un legge donna una}{5} & \rwordcount{capitolo un legge}{5} \wordcount{\tikz[baseline=(node1.base)]\node (node1){una}; \tikz[baseline=(node2.base)]\node (node2){donna};} \begin{tikzpicture}[overlay, remember picture]
             \draw[-latex,draw=red] (node2.north) to[bend right] (node1.north);
             \draw[-latex,draw=red] (node1.south) to[bend right] (node2.south);
        \end{tikzpicture} \\
        & & \textbf{Last-noun  agreement}. Make all determiners agree with the gender of the final noun. & \sout{una} \wordcount{\hlred{un} donna legge un capitolo} & \wordcount{\hlred{una} donna legge un capitolo}  
        \begin{tikzpicture}[overlay, remember picture]
             \draw[black!30, line width=0.5mm] (-19.7, -1.3) -- (2.0, -1.3);
        \end{tikzpicture}\\
        
        & \multirow{5}{*}{Japanese (JP)} & \textbf{Negation}. Insert a negation word at position 4.
        & \begin{CJK}{UTF8}{min}
        \wordcount{女性 は 章 \hlcjk{ない} を 読む}\end{CJK} & \begin{CJK}{UTF8}{min}
        \wordcount{女性 は 章 を 読む \hlcjk{ない}}
        \end{CJK}\\
        & & \textbf{Inversion}. Invert the declarative word order. & \begin{CJK}{UTF8}{min}\rwordcount{読む を 章 は 女性}{5}\end{CJK} & \begin{CJK}{UTF8}{min}\rwordcount{読む を 章}{5}\end{CJK} \wordcount{\tikz[baseline=(node1.base)]\node (node1){\begin{CJK}{UTF8}{min}女性\end{CJK}}; \tikz[baseline=(node2.base)]\node (node2){\begin{CJK}{UTF8}{min}は\end{CJK}};} \begin{tikzpicture}[overlay, remember picture]
             \draw[-latex,draw=red] (node2.north) to[bend right] (node1.north);
             \draw[-latex,draw=red] (node1.south) to[bend right] (node2.south);
        \end{tikzpicture}\\
        & & \textbf{Past tense}. Insert the past tense marker at position 4. & \begin{CJK}{UTF8}{min}\wordcount{女性 は 章 \hlcjk{をた} 読む}\end{CJK} & \begin{CJK}{UTF8}{min}\wordcount{女性 は 章 読む \hlcjk{をた}}\end{CJK} \\
        \bottomrule
        \end{tabular}
    }
    \caption{\textbf{Dataset.} List of grammars, descriptions of the rule defining each grammar, and corresponding positive and negative examples.
    }
    \label{tab:all-template-examples}
\end{table*}

\begin{table*}[!h]
    \centering
    \resizebox{\textwidth}{!}{
        \begin{tabular}{lp{5.5cm}p{6cm}p{7.5cm}}
        \toprule
        & \textbf{Grammar} & \textbf{Positive Example} & \textbf{Negative Example}\\
        \midrule
        \parbox[t]{2mm}{\multirow{3}{*}{\rotatebox[origin=c]{90}{Hierarchical\ }}} & \textbf{Declarative}. Subject, verb, object. & a wug ungos the snorfle & a wug ungos  \tikz[baseline=(node1.base)]\node (node1){snorfle}; \tikz[baseline=(node2.base)]\node (node2){the};
        \begin{tikzpicture}[overlay, remember picture]
             \draw[-latex,draw=red] (node2.north) to[bend right] (node1.north);
             \draw[-latex,draw=red] (node1.south) to[bend right] (node2.south);
        \end{tikzpicture} \\
        & \textbf{Subordinate}. Subject, verb taking a relative clause complement. & Gomu herdles that the blincos stoffle the pelunko & Gomu herdles that the blincos stoffle reads \tikz[baseline=(node1.base)]\node (node1){pelunko}; \tikz[baseline=(node2.base)]\node (node2){the};
        \begin{tikzpicture}[overlay, remember picture]
             \draw[-latex,draw=red] (node2.north) to[bend right] (node1.north);
             \draw[-latex,draw=red] (node1.south) to[bend right] (node2.south);
        \end{tikzpicture}\\
        & \textbf{Passive}. Like \textbf{Declarative}, but in the passive voice. & the gunzle is snugoed by the wugen & the gunzle is snugoed by \tikz[baseline=(node1.base)]\node (node1){wugen}; \tikz[baseline=(node2.base)]\node (node2){the}; \begin{tikzpicture}[overlay, remember picture]
             \draw[-latex,draw=red] (node2.north) to[bend right] (node1.north);
             \draw[-latex,draw=red] (node1.south) to[bend right] (node2.south);
        \end{tikzpicture}\\\midrule%
        \multirow{5}{*}{\rotatebox[origin=c]{90}{Linear\ }} & \textbf{Negation}. Insert ``doesn't'' or ``don't'' at position 5. & \wordcount{the arcuplos ungo a \hlred{doesn't} blorft} &
        \wordcount{the arcuplos ungo a blorft \hlred{doesn't}}\\
        & \textbf{Inversion}. Invert the word order of \textbf{Declarative}. & \rwordcount{snorfle the ungos wug a}{5} & \rwordcount{snorfle the ungos}{5} \wordcount{\tikz[baseline=(node1.base)]\node (node1){a}; \tikz[baseline=(node2.base)]\node (node2){wug};} \begin{tikzpicture}[overlay, remember picture]
             \draw[-latex,draw=red] (node2.north) to[bend right] (node1.north);
             \draw[-latex,draw=red] (node1.south) to[bend right] (node2.south);
        \end{tikzpicture}\\
        & \textbf{Wh-word}. Insert wh-word at position 5. & \wordcount{Did a knurkle gurdles a \hlred{when} skerpo?} & \wordcount{Did a knurkle gurdles a skerpo \hlred{when}?}\\
        \bottomrule
        \end{tabular}
    }
    \caption{\textbf{Jabberwocky dataset.} List of grammars, descriptions of the rules defining each grammar, and positive (grammatical) and negative (ungrammatical) examples for each. We use similar prompt constructions as in the English examples (Also see \S\ref{appendix:prompt}).
    }
    \label{tab:jab-template-examples}
\end{table*}

\section{Experiments}
\subsection{Experiment 1: Few-shot learning accuracy}\label{appendix:expt-1}
\begin{figure*}
    \centering
    \includegraphics[width=\textwidth]{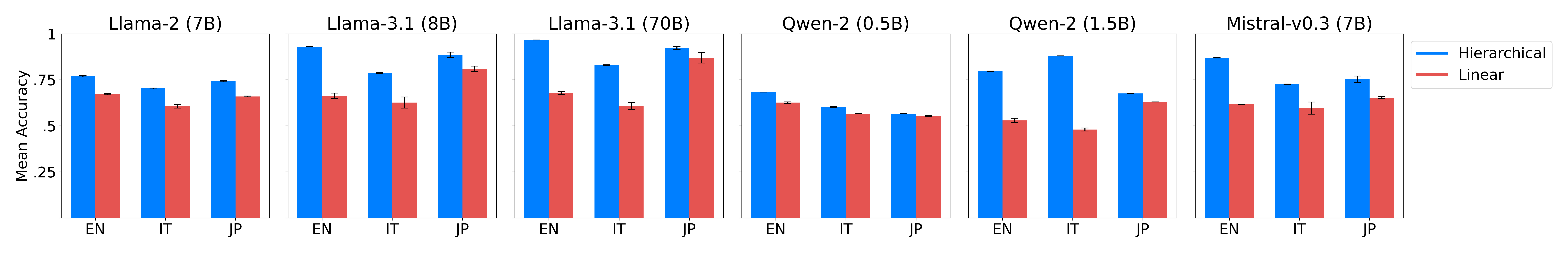}
    \caption{\textbf{Experiment 1.} Model-wise accuracy on the grammaticality judgments task given hierarchical and linear inputs from English, Italian and Japanese(See \S~\ref{sec:exp1} and Tables~\ref{tab:expt1-model-accuracies-conv} and ~\ref{tab:all-template-examples})}
    \label{fig:expt1-model-wise-bars}
\end{figure*}

Experiment 1 (\S\ref{sec:exp1}) assesses the model's performance on grammaticality judgments of hierarchical and linear structures. Here we share statistical comparisons of the accuracy distributions (Table~\ref{tab:expt1-stat-sig}), accuracy values by language (Figure~\ref{fig:expt1-model-wise-bars} and Table~\ref{tab:expt1-model-accuracies-conv}) and grammar-wise accuracy values (Table~\ref{tab:expt1-model-accuracies-conv}).
\subsubsection{Example Prompts}
\label{appendix:prompt}
We present example prompts from one of the hierarchical structures for each language. The prompt skeleton is in English, regardless of the language used for the examples. We intentionally strip the final whitespace after \texttt{A:}, as the model expects a leading space within the answer token (and thus, it should not be present in the prompt).

\paragraph{English example.}

\texttt{"Here are English sentences that either follow or break a grammar rule. Each sentence is labeled 'Yes' if it follows the rule and 'No' if it doesn't. Label the final sentence as 'Yes' or 'No' based on whether it follows the same rule.
\newline
\newline
Q: Is this sentence grammatical? Yes or No: a woman drinks espresso the \newline
A: No \newline
\newline
Q: Is this sentence grammatical? Yes or No: the architects touch a mouse \newline
A: Yes \newline
\newline
Q: Is this sentence grammatical? Yes or No: the women eat cucumber the \newline
A: No \newline
\newline
Q: Is this sentence grammatical? Yes or No: the writers drink a lemonade \newline
A: Yes \newline
\newline
Q: Is this sentence grammatical? Yes or No: a teacher touches a lightbulb \newline
A: Yes \newline
\newline
Q: Is this sentence grammatical? Yes or No: the actress touches toy a \newline
A: No \newline
\newline
Q: Is this sentence grammatical? Yes or No: a boy kicks bottle a \newline
A: No \newline
\newline
Q: Is this sentence grammatical? Yes or No: the woman pushes toy a \newline
A: No \newline
\newline
Q: Is this sentence grammatical? Yes or No: a professor reads a poem \newline
A: Yes \newline
\newline
Q: Is this sentence grammatical? Yes or No: the orators read a story \newline
A: Yes \newline
\newline
Q: Is this sentence grammatical? Yes or No: the doctor drinks milkshake the \newline
A:" \newline}

\paragraph{Italian example.}

\texttt{"Here are Italian sentences that either follow or break a grammar rule. Each sentence is labeled 'Yes' if it follows the rule and 'No' if it doesn't. Label the final sentence as 'Yes' or 'No' based on whether it follows the same rule.
\newline
\newline
Q: Is this sentence grammatical? Yes or No: una donna beve espresso il \newline
A: No \newline
\newline
Q: Is this sentence grammatical? Yes or No: l' architette toccano il topo \newline
A: Yes \newline
\newline
Q: Is this sentence grammatical? Yes or No: le donne mangiano cetriolo il \newline
A: No \newline
\newline
Q: Is this sentence grammatical? Yes or No: le scrittrici bevono la limonata \newline
A: Yes \newline
\newline
Q: Is this sentence grammatical? Yes or No: un' insegnante tocca una lampadina \newline
A: Yes \newline
\newline
Q: Is this sentence grammatical? Yes or No: l attrice tocca giocattolo un \newline
A: No \newline
\newline
Q: Is this sentence grammatical? Yes or No: un ragazzo calcia bottiglia una \newline
A: No \newline
\newline
Q: Is this sentence grammatical? Yes or No: la donna spinge giocattolo un \newline
A: No \newline
\newline
Q: Is this sentence grammatical? Yes or No: una professoressa legge un poema \newline
A: Yes \newline
\newline
Q: Is this sentence grammatical? Yes or No: gli oratori leggono la storia \newline
A: Yes \newline
\newline
Q: Is this sentence grammatical? Yes or No: la dottoressa beve frappè il \newline
A:"}

\paragraph{Japanese example.}
\texttt{"Here are Japanese sentences that either follow or break a grammar rule. Each sentence is labeled 'Yes' if it follows the rule and 'No' if it doesn't. Label the final sentence as 'Yes' or 'No' based on whether it follows the same rule.
\newline
\newline
Q: Is this sentence grammatical? Yes or No: \begin{CJK}{UTF8}{min}女性はエスプレッソ飲むを\end{CJK} \newline
A: No \newline
\newline
Q: Is this sentence grammatical? Yes or No: \begin{CJK}{UTF8}{min}建築家たちはマウスを触る\end{CJK} \newline
A: Yes \newline
\newline
Q: Is this sentence grammatical? Yes or No: \begin{CJK}{UTF8}{min}女性たちは胡瓜食べるを\end{CJK} \newline
A: No \newline
\newline
Q: Is this sentence grammatical? Yes or No: \begin{CJK}{UTF8}{min}作家たちはレモネードを飲む\end{CJK} \newline
A: Yes \newline
\newline
Q: Is this sentence grammatical? Yes or No: \begin{CJK}{UTF8}{min}教師は電球を触る\end{CJK} \newline
A: Yes \newline
\newline
Q: Is this sentence grammatical? Yes or No: \begin{CJK}{UTF8}{min}女優は玩具触るを\end{CJK} \newline
A: No \newline
\newline
Q: Is this sentence grammatical? Yes or No: \begin{CJK}{UTF8}{min}少年はボトル蹴るを\end{CJK} \newline
A: No \newline
\newline
Q: Is this sentence grammatical? Yes or No: \begin{CJK}{UTF8}{min}女性は玩具押すを\end{CJK} \newline
A: No \newline
\newline
Q: Is this sentence grammatical? Yes or No: \begin{CJK}{UTF8}{min}教授は詩を読む\end{CJK} \newline
A: Yes \newline
\newline
Q: Is this sentence grammatical? Yes or No: \begin{CJK}{UTF8}{min}演説家たちは小説を読む\end{CJK} \newline
A: Yes \newline
\newline
Q: Is this sentence grammatical? Yes or No: \begin{CJK}{UTF8}{min}医者はミルクセーキ飲むを\end{CJK} \newline
A:"}

\begin{figure*}
    \centering
    \includegraphics[width=\textwidth]{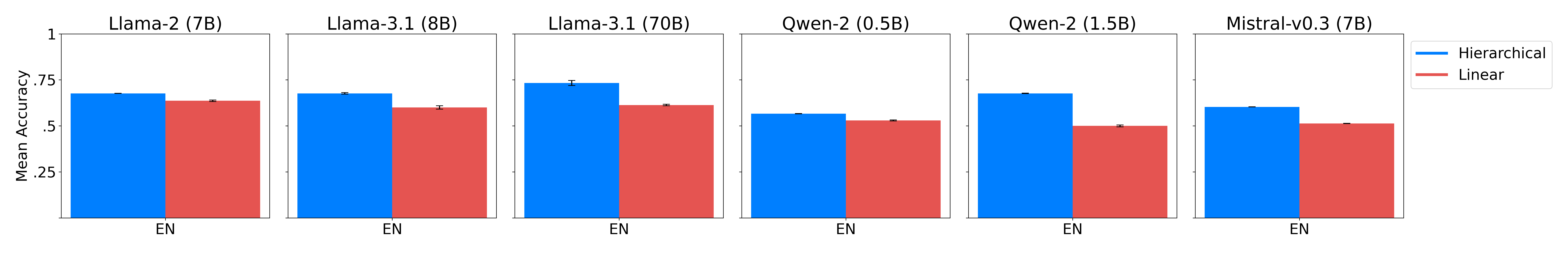}
    \caption{\textbf{Experiment 1.} Model-wise accuracy on the grammaticality judgments task given hierarchical and linear Jabberwocky inputs (See \S~\ref{sec:exp1} and Tables~\ref{tab:expt1-model-accuracies-nonce} and ~\ref{tab:jab-template-examples})}
    \label{fig:expt1-model-wise-bars-nonce}
\end{figure*}

\begin{table*}[ht]
    \centering
    \begin{tabular}{lll}
    \toprule
    Language   &   Test-Statistic &   P-value \\
    \midrule
    EN         &   5.92   &   p $<$ 0.001 \\
    IT         &   271.5  &   p $<$ 0.001 \\
    JP         &   203    &   0.2          \\
    \bottomrule
    \end{tabular}
    \caption{\textbf{Experiment 1.} Results from a Mann-Whitney U-Test testing if the model accuracy on the grammaticality judgment task of hierarchical inputs is significantly different from that on linear inputs, when including English, Italian, and Japanese hierarchical and linear inputs (p $<$ 0.05).}
    \label{tab:expt1-stat-sig}
\end{table*}

\begin{table*}[ht]
\centering
\resizebox{\textwidth}{!}{%
\begin{tabular}{lccccccc}
\toprule
Grammar & Llama-2-7B & Llama-3.1-8B & Llama-3.1-70B & Qwen-2-0.5B & Qwen-2-1.5B & Mistral-v0.3 \\
\midrule
EN Declarative (H)    & 0.80 & 0.94 & 0.96 & 0.68 & 0.85 & 0.91 \\
EN Subordinate (H)    & 0.68 & 0.92 & 0.98 & 0.67 & 0.76 & 0.83 \\
EN Passive (H)        & 0.83 & 0.93 & 0.96 & 0.70 & 0.78 & 0.87 \\
EN Negation (L)       & 0.76 & 0.83 & 0.81 & 0.65 & 0.39 & 0.60 \\
EN Inversion (L)      & 0.65 & 0.55 & 0.61 & 0.69 & 0.65 & 0.62 \\
EN Wh-word (L)        & 0.61 & 0.61 & 0.62 & 0.54 & 0.55 & 0.63 \\
\midrule
IT Declarative (H)    & 0.74 & 0.81 & 0.89 & 0.63 & 0.90 & 0.76 \\
IT Subordinate (H)    & 0.64 & 0.71 & 0.78 & 0.52 & 0.87 & 0.69 \\
IT Passive (H)        & 0.73 & 0.84 & 0.82 & 0.66 & 0.87 & 0.73 \\
IT Negation (L)       & 0.73 & 0.87 & 0.80 & 0.60 & 0.60 & 0.85 \\
IT Inversion (L)      & 0.61 & 0.53 & 0.52 & 0.59 & 0.46 & 0.50 \\
IT Gender Agreement (L) & 0.48 & 0.48 & 0.50 & 0.51 & 0.38 & 0.44 \\
\midrule
JP Declarative (H)    & 0.72 & 0.95 & 0.99 & 0.54 & 0.67 & 0.78 \\
JP Subordinate (H)    & 0.68 & 0.72 & 0.80 & 0.57 & 0.65 & 0.58 \\
JP Passive (H)        & 0.83 & 0.99 & 0.98 & 0.59 & 0.71 & 0.90 \\
JP Negation (L)       & 0.63 & 0.94 & 0.99 & 0.61 & 0.64 & 0.75 \\
JP Inversion (L)      & 0.62 & 0.65 & 0.63 & 0.55 & 0.61 & 0.63 \\
JP Past-tense (L)     & 0.73 & 0.84 & 0.99 & 0.50 & 0.64 & 0.58 \\
\bottomrule
\end{tabular}
}
\caption{\textbf{Experiment 1.} Model accuracies on the grammaticality judgment task for English, Italian, and Japanese hierarchical and linear inputs.}
\label{tab:expt1-model-accuracies-conv}
\end{table*}

\begin{table*}[ht]
\centering
\resizebox{\textwidth}{!}{%
\begin{tabular}{lcccccc}
\toprule
Grammar & Llama-2-7B & Llama-3.1-8B & Llama-3.1-70B & Qwen-2-0.5B & Qwen-2-1.5B & Mistral-v0.3 \\
\midrule
Declarative (H) & 0.68 & 0.64 & 0.70 & 0.57 & 0.69 & 0.61 \\
Subordinate (H) & 0.64 & 0.62 & 0.61 & 0.53 & 0.63 & 0.58 \\
Passive (H)     & 0.71 & 0.77 & 0.89 & 0.60 & 0.71 & 0.62 \\
Negation (L)    & 0.70 & 0.73 & 0.70 & 0.54 & 0.45 & 0.54 \\
Inversion (L)   & 0.66 & 0.57 & 0.59 & 0.59 & 0.60 & 0.52 \\
Wh-word (L)     & 0.55 & 0.50 & 0.55 & 0.46 & 0.45 & 0.48 \\
\bottomrule
\end{tabular}%
}
\caption{\textbf{Experiment 1.} Model accuracies on the grammaticality judgment task for Jabberwocky grammars.}
\label{tab:expt1-model-accuracies-nonce}
\end{table*}

\subsection{Experiment 2: Identify MLP and Attention Components with the highest $\hat{\text{IE}}$}\label{appendix:exp2}

\begin{figure*}
    \centering
    \begin{subfigure}[b]{\textwidth}
        \centering
        \includegraphics[trim={1cm 0 1cm 0}, clip, width=\textwidth]{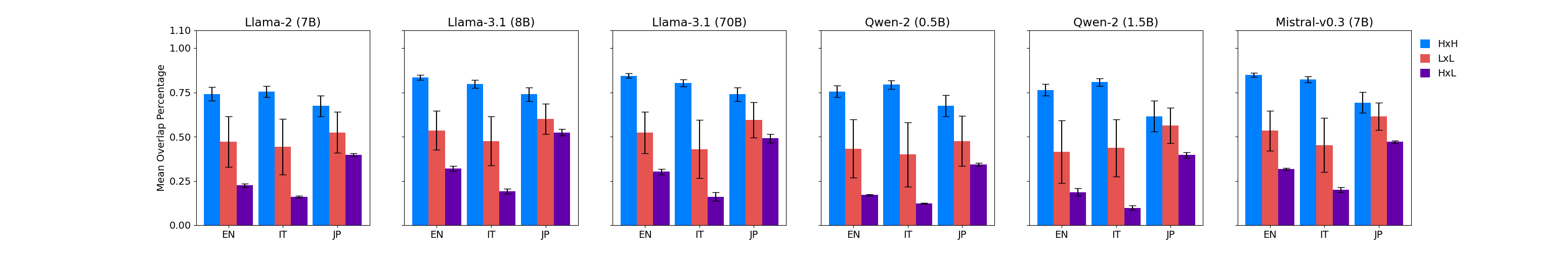}
        \caption{English, Italian, and Japanese grammars}
        \label{fig:expt2-model-wise-overlaps-en-it-ja}
    \end{subfigure}
    \begin{subfigure}[b]{\textwidth}
        \centering
        \includegraphics[trim={1cm 0 1cm 0}, clip, width=\textwidth]{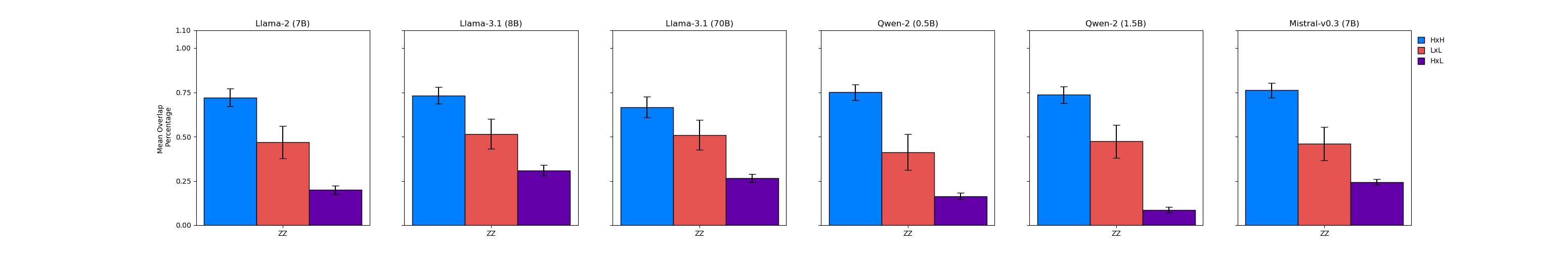}
        \caption{Jabberwocky grammars.}
        \label{fig:expt2-model-wise-overlaps-jab}
    \end{subfigure}
    
    \caption{\textbf{Experiment 2.} Mean pairwise neuron overlaps, by model, for the top 1\% of MLP and attention neurons by $\hat{IE}$ between hierarchical and linear inputs. (See \S~\ref{sec:exp2})}
    \label{fig:expt2-model-overlaps-all}
\end{figure*}

\begin{figure*}[]
    \includegraphics[width=\textwidth]{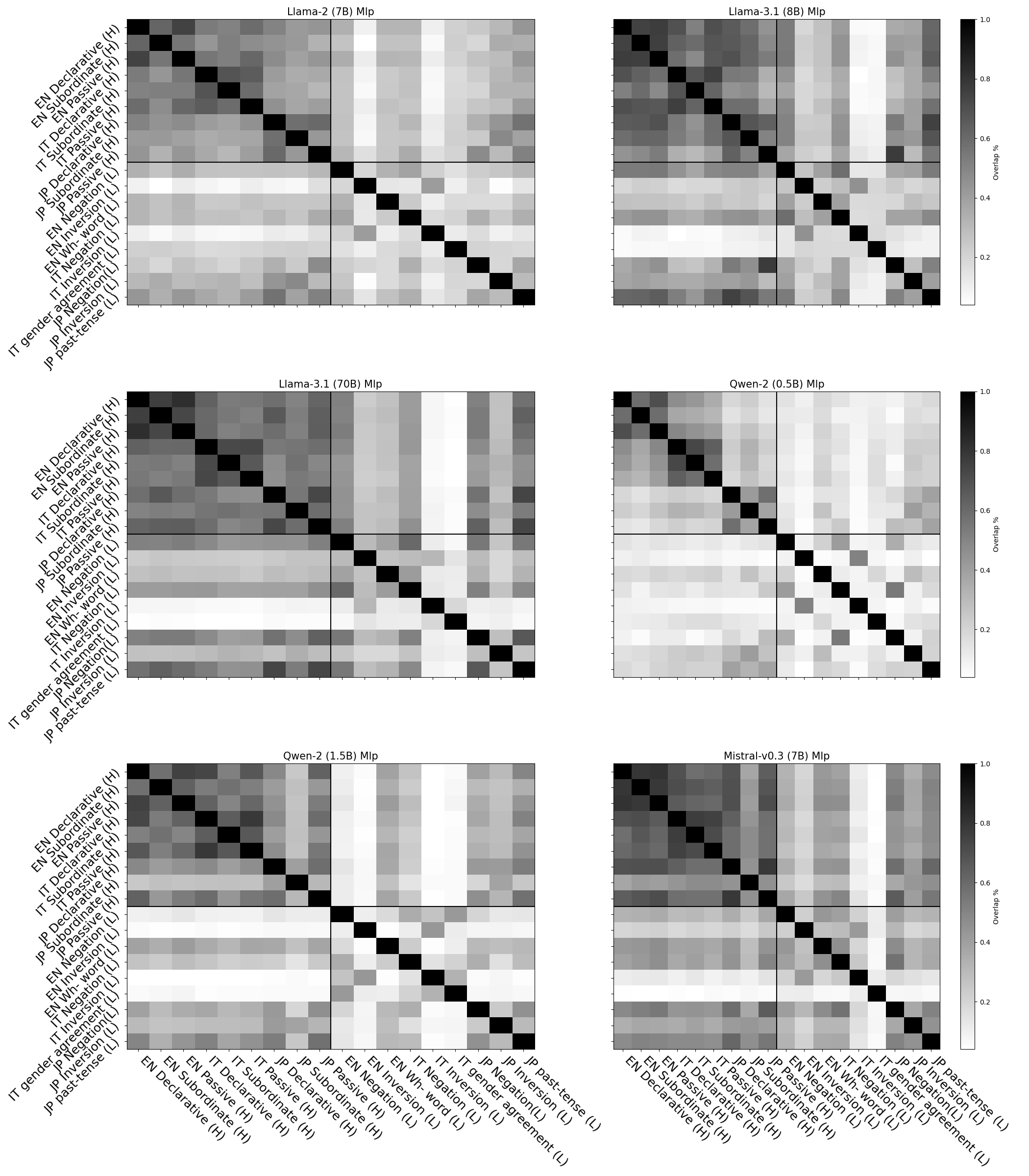}
    \caption{\textbf{Experiment 2.} MLP neuron overlaps by model for English, Italian and Japanese grammars.}
    \label{fig:expt-2-conv-comp-mlp}  
\end{figure*}

\begin{figure*}[]
    \centering
    \includegraphics[width=\textwidth]{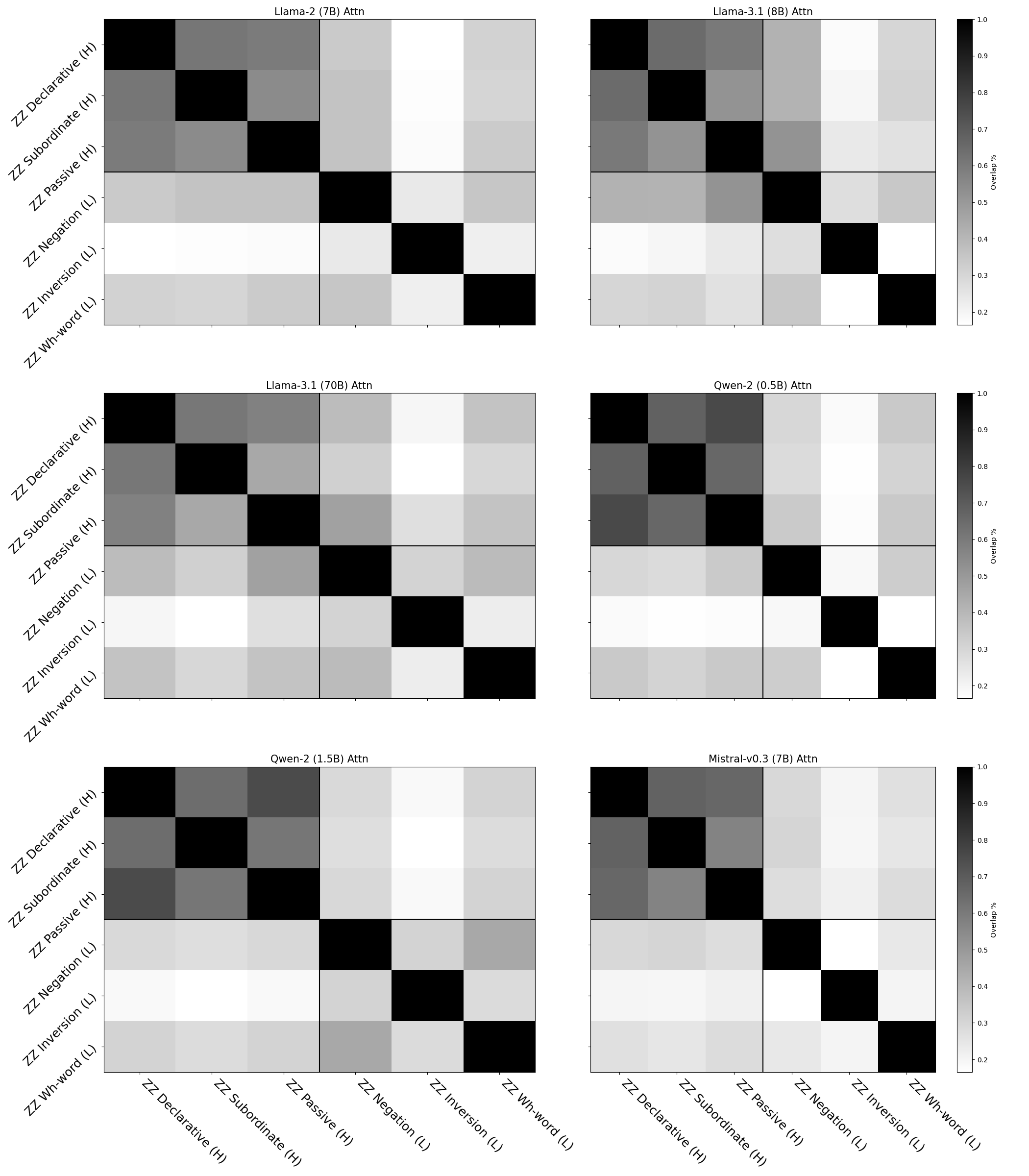}
    \caption{\textbf{Experiment 2.} Attention Overlaps by model for Jabberwocky grammars.}
    \label{fig:expt-2-conv-comp-attn}  
\end{figure*}

Experiment 2 locates MLP and Attention neurons that are implicated in processing hierarchical and linear structures, and investigates if these components are disjoint. The language-level pairwise overlaps for all models across English, Italian and Japanese hierarchical and linear inputs, as well as Jabberwocky hierarchical and linear inputs is given in Figures ~\ref{fig:expt2-model-overlaps-all}. Grammar wise overlaps for MLP and attention components for English, Italian, and Japanese grammars are shown in Figures~\ref{fig:expt-2-conv-comp-mlp} and ~\ref{fig:expt-2-conv-comp-attn}, respectively.  Grammar wise overlaps for MLP and attention components for Jabberwocky grammars are shown in Figures~\ref{fig:expt-2-conv-comp-mlp} and ~\ref{fig:expt-2-conv-comp-attn}, respectively. We also share results testing whether the pairwise mean overlaps of H, L, and HxL structures are significantly different among grammars using English, Italian, and Japanese versus Jabberwocky tokens in Tables~\ref{tab:expt2-stat-sig-conv} and \ref{tab:expt2-stat-sig-jab}.

\begin{figure*}[]
    \centering
    \includegraphics[width=\textwidth]{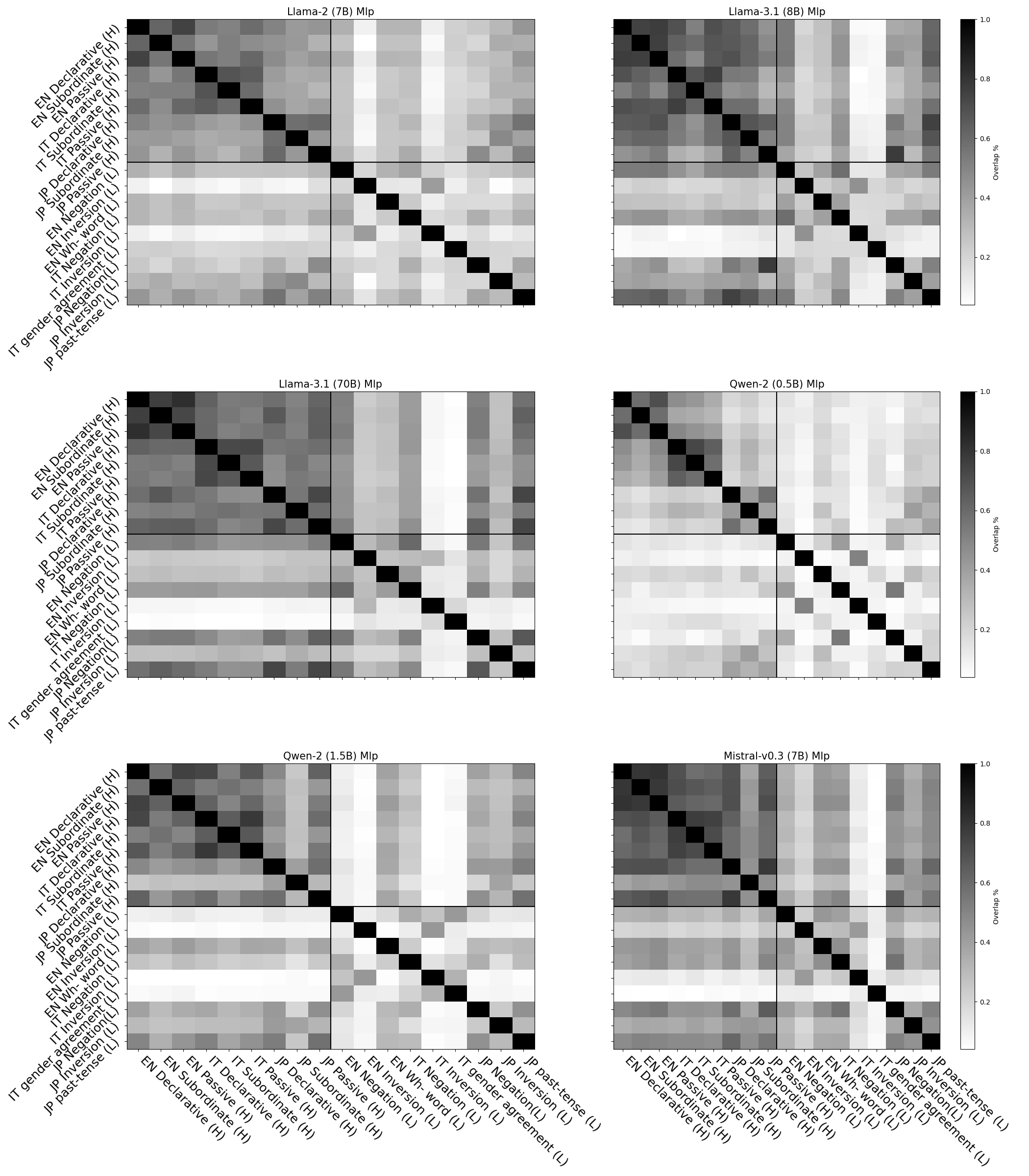}
    \caption{\textbf{Experiment 2.} Attention neuron overlaps by model for English, Italian and Japanese grammars.}
    \label{fig:expt-2-jab-comp-attn}  
\end{figure*}

\begin{figure*}[]
    \includegraphics[width=\textwidth]{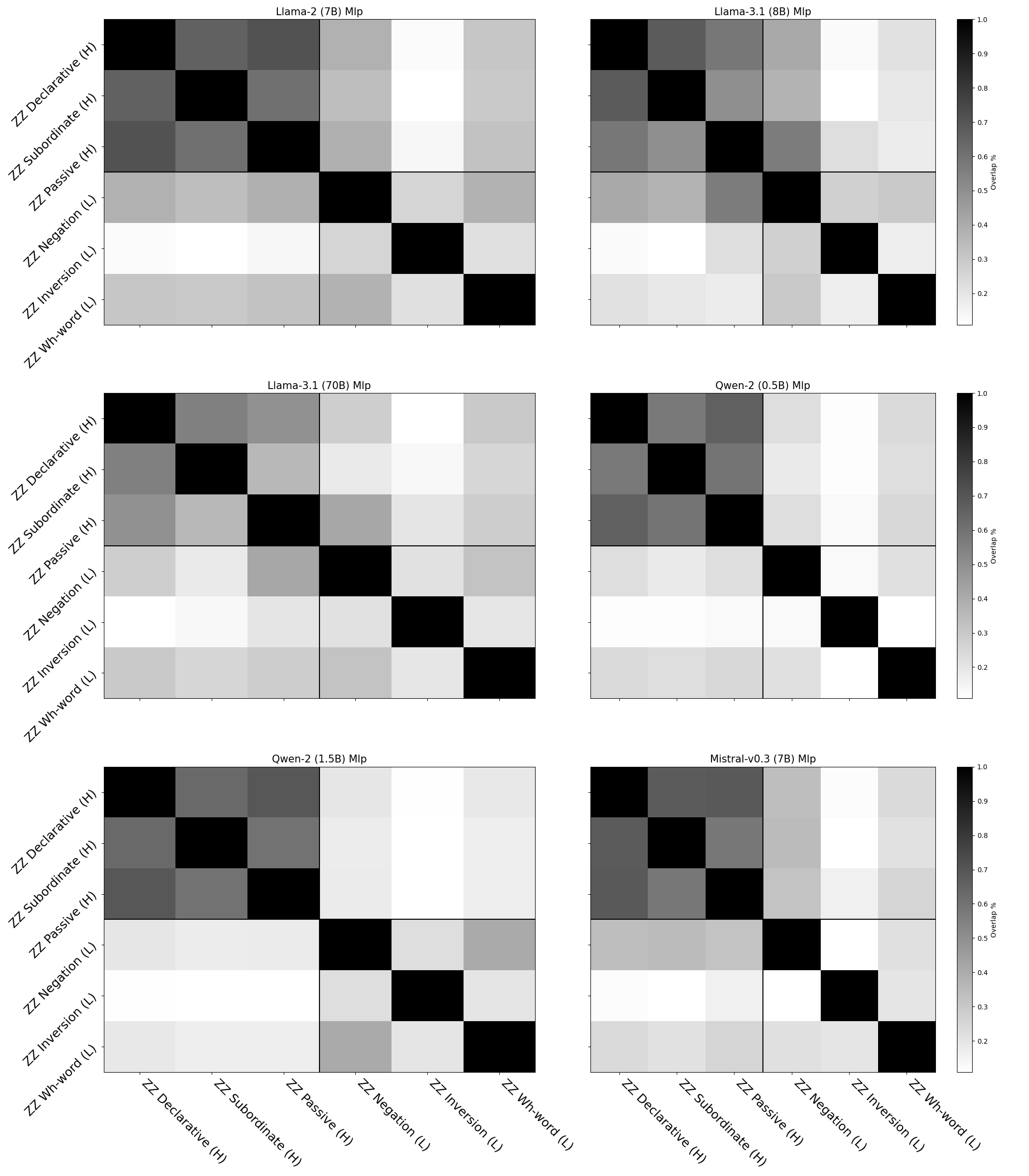}
    \caption{\textbf{Experiment 2.} MLP neuron overlaps by model for Jabberwocky grammars.}
    \label{fig:expt-2-jab-comp-mlp}  
\end{figure*}

\begin{table*}[!ht]
\centering
\begin{tabular}{lcccccc}
\toprule
Components & (Test-Statistic, P-value) \\
\midrule
 H-H vs L-L & (8424, p < 0.001) \\
 H-H vs H-L & (11594, p < 0.001) \\
 L-L vs H-L & (7792, p < 0.001)\\
\bottomrule
\end{tabular}%
\caption{\textbf{Experiment 2.} Results from a Mann-Whitney U-Test investigating whether the overlap percentages for different components across 7 models is significantly different for Jabberwocky grammars. We compare distributions of the mean overlap percentages for the top 1\% of MLP and attention components (p < 0.05, N = 108).}
\label{tab:expt2-stat-sig-jab}
\end{table*}

\begin{table*}[!ht]
\centering
\begin{tabular}{lccccc}
\toprule
Language & Components & Test Statistic & P-value \\
\midrule
\multirow{3}{*}{English} 
 & H-H vs L-L & 74794.0 & $p < 0.001$ \\
 & H-H vs H-L & 101160.0 & $p < 0.001$ \\
 & L-L vs H-L & 65649.0 & $p < 0.001$ \\
\midrule
\multirow{3}{*}{Italian} 
 & H-H vs L-L & 74794.0 & $p < 0.001$ \\
 & H-H vs H-L & 101160.0 & $p < 0.001$ \\
 & L-L vs H-L & 65649.0 & $p < 0.001$ \\
\midrule
\multirow{3}{*}{Japanese} 
 & H-H vs L-L & 74794.0 & $p < 0.001$ \\
 & H-H vs H-L & 101160.0 & $p < 0.001$ \\
 & L-L vs H-L & 65649.0 & $p < 0.001$ \\
\bottomrule
\end{tabular}%
\caption{\textbf{Experiment 2.} Results from a Mann-Whitney U-Test investigating whether the overlap percentages for different components across 7 models are significantly different for English, Italian, and Japanese grammars. We compare distributions of the mean overlap percentages for the top 1\% of MLP and attention components (p < 0.05, N = 108).}
\label{tab:expt2-stat-sig-conv}
\end{table*}

\subsection{Experiment 3: Ablations of top 1\% of Attention and MLP Components}\label{appendix:exp3}

\begin{figure*}
    \centering
    \begin{subfigure}[b]{\textwidth}
        \centering
        \includegraphics[trim={1cm 0 1cm 0}, clip, width=\textwidth, scale=0.55]{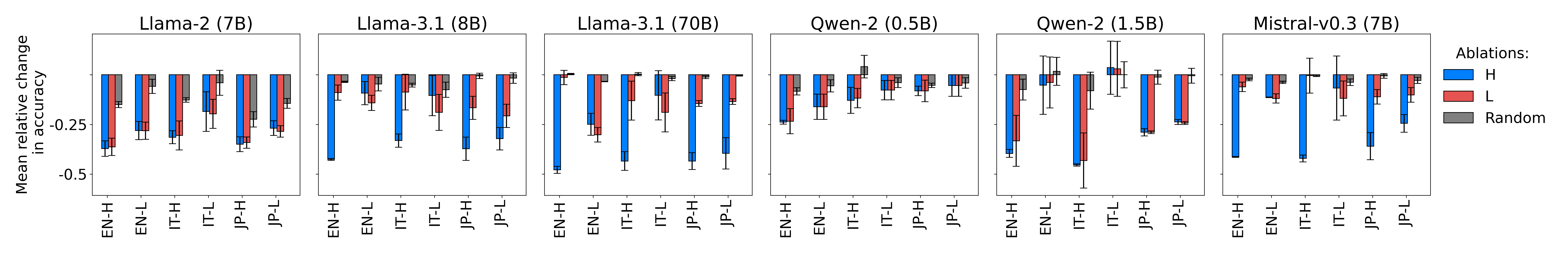}
        \caption{English, Italian, and Japanese grammars}
        \label{fig:expt3-model-ablations-en-it-ja}
    \end{subfigure}
    \begin{subfigure}[b]{\textwidth}
        \centering
        \includegraphics[trim={1cm 0 1cm 0}, clip, width=\textwidth, scale=0.55]{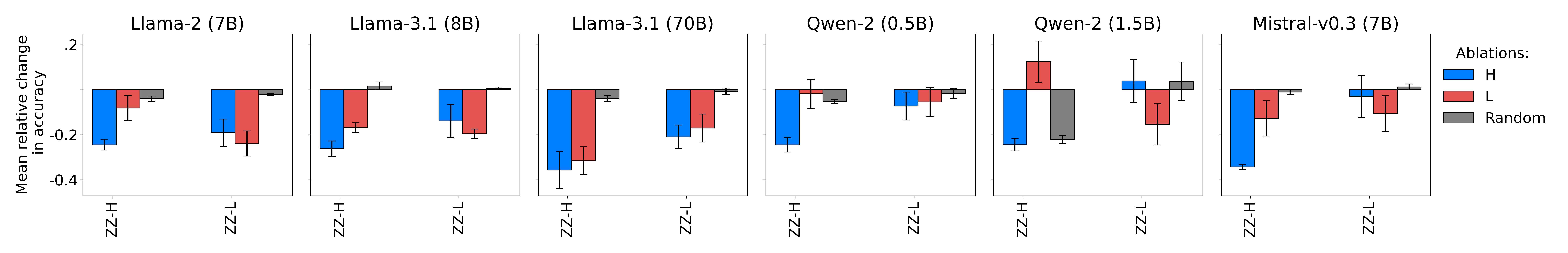}
        \caption{Jabberwocky grammars.}
        \label{fig:expt3-model-ablations-jab}
    \end{subfigure}
    
    \caption{\textbf{Experiment 3.} Mean relative change in accuracy by model, when ablating the top 1\% of attention and MLP neurons by $\hat{IE}$ between hierarchical and linear inputs. (See \S~\ref{sec:exp3})}
    \label{fig:expt3-model-ablations-all}
\end{figure*}

Experiment 3 considers selective ablations of hierarchy and linearity sensitive components, and evaluates how these ablations impact the accuracy of the model on the in-context learning task. We share ablation results by model in Figure~\ref{fig:expt3-model-ablations-en-it-ja} for English, Italian, and Japanese grammars. Through model-wise comparisons, we find that relative accuracy decreases on hierarchical grammars are significantly different for English, Italian and Japanese grammars depending on whether  hierarchical, linear, or uniformly sampled components are ablated (see Table~\ref{tab:exp3-stat-sig-conv}). However, the same is not true for linear grammars---relative accuracy decreases are not significantly different between ablations of hierarchical/linear components. In the case of Italian structures, ablating hierarchy-sensitive components appears akin to ablating uniformly sampled components.

Additionally, we also run ablation experiments on jabberwocky grammars using components that are sensitive to hierarchical or linear jabberwocky grammars. We share ablation results by model for jabberwocky grammars in Figure~\ref{fig:expt3-model-ablations-jab}. Here also we find that ablating hierarchy versus linearity sensitive components can cause a significant difference in the decrease in accuracy on jabberwocky hierarchical grammars relative to the no ablation case. This is not true for jabberwocky linear grammars (See Table~\ref{tab:exp3-stat-sig-nonce}).

\begin{table*}[]
    \centering
    \begin{tabular}{ccccc}
    \toprule
    \textbf{Ablation on} & \textbf{Ablation comparisons (Top 1\% of components)} & \textbf{Test-statistic} & \textbf{P-Value} \\ 
    \midrule
    
    EN (H) & H-components vs L-components & 31.5 & < 0.001 \\
    EN (H) & H-components vs R-components & 0.0  & < 0.001 \\
    EN (H) & L-components vs R-components & 76.5 & 0.01  \\
    EN (L) & H-components vs L-components & 173.5 & 0.73 \\
    EN (L) & H-components vs R-components & 52.5 & < 0.001 \\
    EN (L) & L-components vs R-components & 40.0 & < 0.001 \\
    
    IT (H) & H-components vs L-components & 65.0 & < 0.001 \\
    IT (H) & H-components vs R-components & 15.5 & < 0.001 \\
    IT (H) & L-components vs R-components & 76.5 & 0.01 \\
    IT (L) & H-components vs L-components & 193.0 & 0.33 \\
    IT (L) & H-components vs R-components & 155.5 & 0.85 \\
    IT (L) & L-components vs R-components & 91.5 & 0.03 \\
    
    JP (H) & H-components  vs L-components  & 77.0 & 0.01 \\
    JP (H) & H-components  vs R-components  & 17.0 & < 0.001 \\
    JP (H) & L-components  vs R-components  & 41.0 & < 0.001 \\
    JP (L) & H-components  vs L-components  & 102.0 & 0.06 \\
    JP (L) & H-components  vs R-components  & 25.0 & < 0.001 \\
    JP (L) & L-components  vs R-components  & 43.5 & < 0.001 \\
    
    \bottomrule
    \end{tabular}
    \caption{\textbf{Experiment 3.} Results from a Mann-Whitney U-test investigating whether ablating uniformly sampled model components, as well as components used in the grammaticality judgment tasks of hierarchical and linear grammars containing natural language lexicons significantly differ in how they suppress performance on English, Italian, and Japanese hierarchical and linear grammars. Mean ablations are applied from the structure being judged in the task. }
    \label{tab:exp3-stat-sig-conv}
    \end{table*}

\begin{table*}[]
    \centering
    \begin{tabular}{ccccc}
    \toprule
    \textbf{Language} & \textbf{Ablation} & \textbf{Test-statistic} & \textbf{P-Value} \\ 
    \midrule
    
    ZZ (H) & ZZ(H) vs ZZ(L) & 33.0 & < 0.001 \\
    ZZ (H) & ZZ(H) vs ZZ(R) & 14.5 & < 0.001 \\
    ZZ (H) & ZZ(L) vs ZZ(R) & 131.5 & 0.34 \\
    ZZ (L) & ZZ(H) vs ZZ(L) & 197.0 & 0.27 \\
    ZZ (L) & ZZ(H) vs ZZ(R) & 84.5 & 0.01 \\
    ZZ (L) & ZZ(L) vs ZZ(R) & 44.5 & < 0.001 \\
    
    \bottomrule
    \end{tabular}
    \caption{\textbf{Experiment 3.} Results from a Mann-Whitney U-test investigating whether ablating uniformly sampled model components, as well as components used in the grammaticality judgment tasks of jabberwocky hierarchical and linear grammars significantly differ in how they suppress performance on jabberwocky hierarchical and linear grammars. Mean ablations are applied from the structure being judged in the task.}
    \label{tab:exp3-stat-sig-nonce}
    \end{table*}

\subsection{Experiment 4: Are neurons identified in experiment 3 sensitive to hierarchical structure or in-distribution lexical tokens?}\label{appendix:exp4}

\begin{table*}[]
\centering
\begin{tabular}{cccc}
\toprule
\textbf{Ablation} & \textbf{Structure Type} & \textbf{Test-statistic} & \textbf{P-Value} \\ 
\midrule
EN (H) vs EN (L) & ZZ (H) & 69.5  & 0.004   \\
EN (H) vs EN (Random) & ZZ (H) & 11.0  & < 0.001 \\
EN (L) vs EN (Random) & ZZ (H) & 68.5  & 0.003   \\
\midrule
EN (H) vs EN (L) & ZZ (L) & 188.5 & 0.41   \\
EN (H) vs EN (Random) & ZZ (L) & 104.0 & 0.07   \\
EN (L) vs EN (Random) & ZZ (L) & 87.5  & 0.02   \\
\bottomrule
\end{tabular}
\caption{\textbf{Experiment 4.} Results from a Mann-Whitney U-test investigating whether ablating uniformly sampled model components, as well as components used in the grammaticality judgment tasks of hierarchical and linear English grammars significantly differ in how they suppress performance on jabberwocky hierarchical and linear grammars. }
\label{tab:exp4-stat-sig-ablations}
\end{table*}

\begin{figure*}
    \centering
    \includegraphics[trim={1cm 0 1cm 0}, clip, width=\textwidth, scale=0.55]{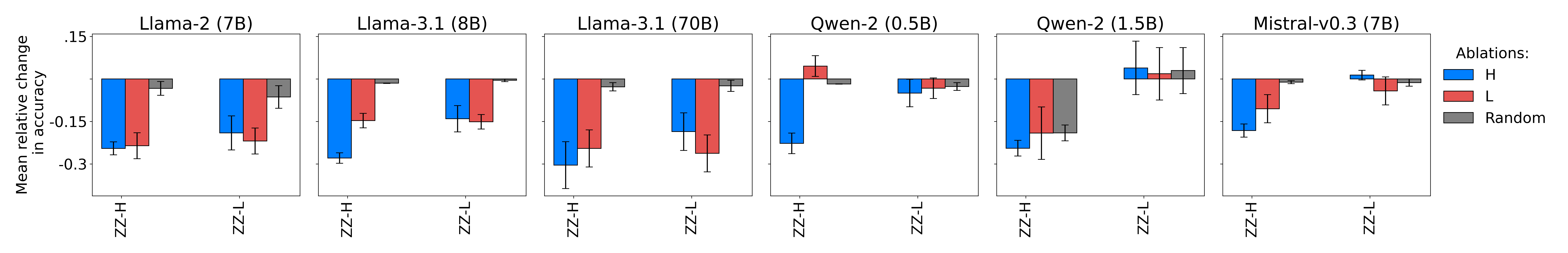}
    \caption{\textbf{Experiment 4.} Mean relative change in accuracy by model, when ablating the top 1\% of attention and MLP neurons pertaining to English hierarchical and linear grammars. Model is tested on Jabberwocky grammars post ablation, and performance decrease is measured on hierarchical and linear inputs. (See \S~\ref{sec:expt-4-jabberwocky})}
    \label{fig:expt4-model-ablations}
\end{figure*}

\begin{figure*}[]
    \includegraphics[width=\textwidth]{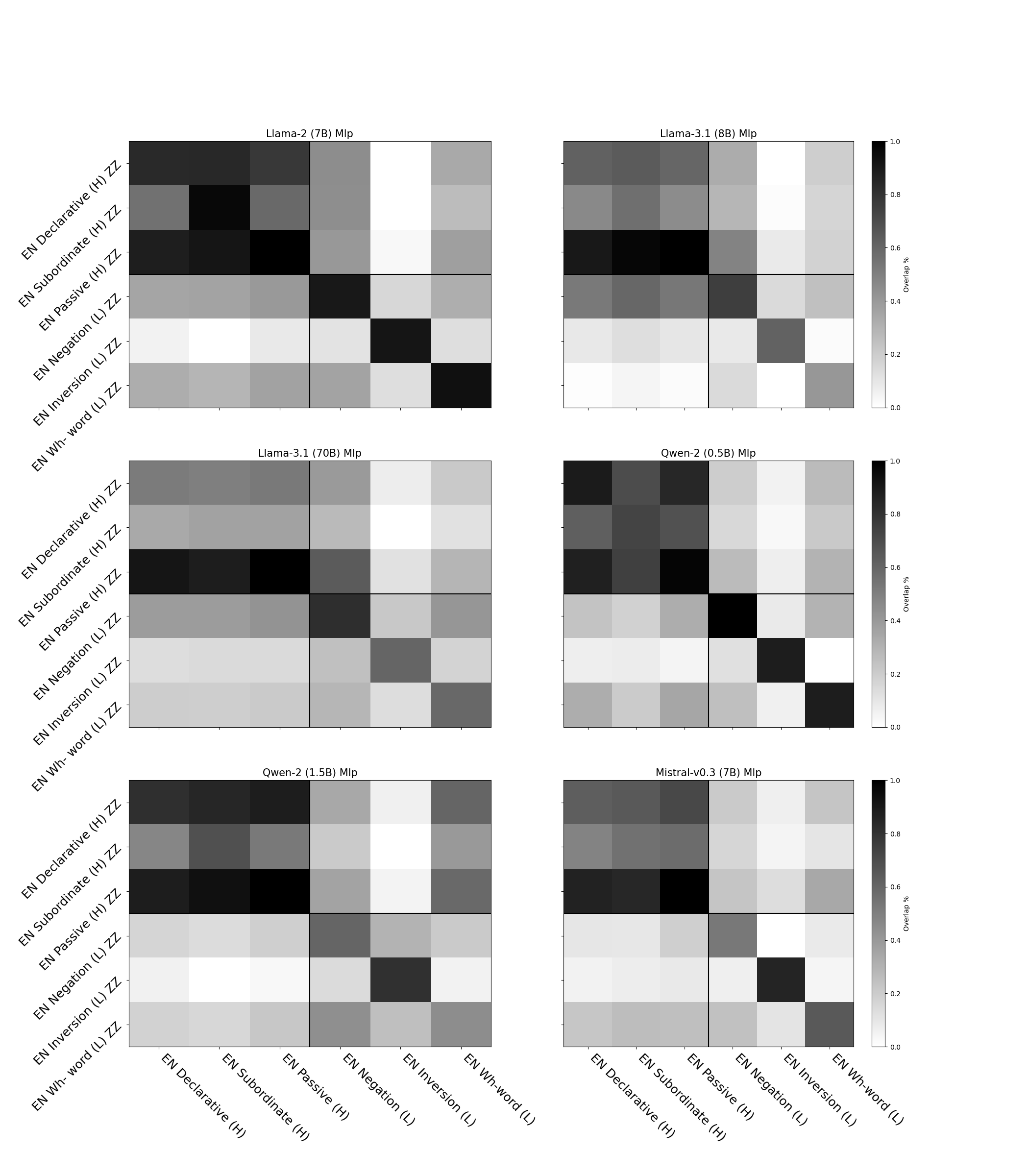}
    \caption{\textbf{Experiment 4.} MLP Overlaps by model between English and Jabberwocky grammars}
    \label{fig:expt4-jab-comp-mlp}  
\end{figure*}

\begin{figure*}[]
    \centering
    \includegraphics[width=\textwidth]{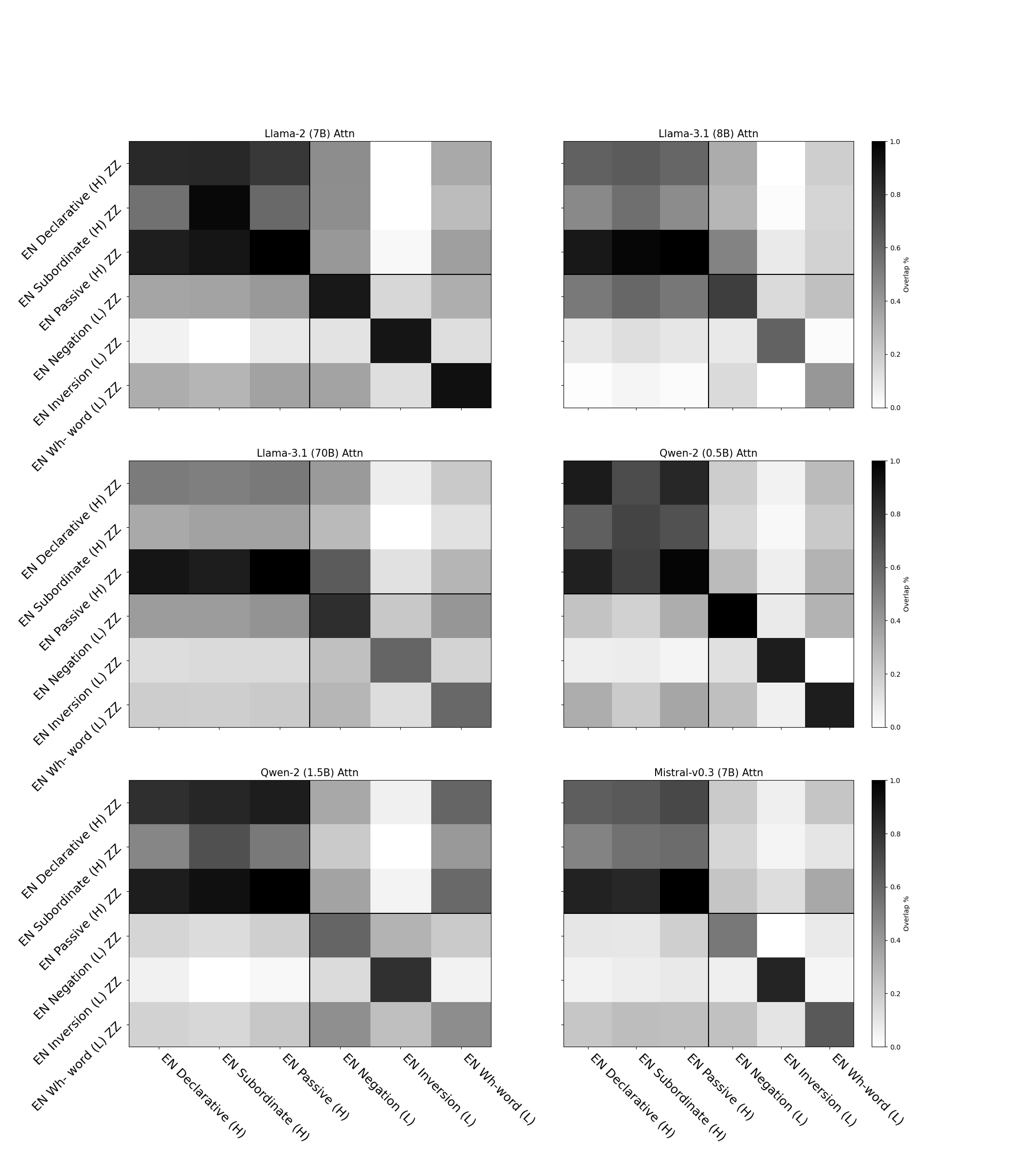}
    \caption{\textbf{Experiment 4.} Attention Overlaps by model between English and Jabberwocky grammars}
    \label{fig:expt4-jab-comp-attn}  
\end{figure*}

Experiment 4 considers selective ablations of the top 1\% of hierarchy and linearity sensitive components, and evaluates how these ablations impact the accuracy of the model on the in-context learning task, when processing Jabberwocky grammars. If the neurons discovered in Experiment 3 are \textit{not} sensitive to hierarchical structure and instead sensitive to in-distribution tokens, these ablations should not cause a decrease in model performance on Jabberwocky grammars which are composed of meaningless words. Alternatively, any decreases in model performance on Jabberwocky grammars should be caused by neurons in the $L$ set which are, say, sensitive to out of distribution inputs. Ablation results by model are in Figure~\ref{fig:expt4-model-ablations}. We also present grammar-wise overlaps of the top 1\% of attention and MLP neurons for hierarchical and linear English and Jabberwocky grammars in Figures~\ref{fig:expt4-jab-comp-attn} and \ref{fig:expt4-jab-comp-mlp} respectively, and show that the difference in overlaps between these grammars is statistically significant in Table~\ref{tab:exp4-stat-sig-ablations}. Then, we test the relative change in accuracy in Jabberwocky grammars after ablating components sensitive to hierarchical and linear English structures as well as uniformly sampled components. Ablating hierarchy vs.\ linearity sensitive components that are sensitive to the English task, causes a significantly different decrease in model performance on Jabberwocky hierarchical grammars. However, the same is not true for linear Jabberwocky grammars where ablating hierarchy sensitive components is no different from ablating linearity-sensitive or uniformly sampled components (see Table~\ref{tab:expt4-stat-sig-jab}. This suggests that the components identified in Experiment 3 are at least partially sensitive to the structure of the inputs.

\begin{table*}[!ht]
\centering
\begin{tabular}{lcccccc}
\toprule
Components & (Test-Statistic, P-value) \\
\midrule
 H(ZZ x EN) vs L(ZZ x EN) & (9678, p < 0.001) \\
 H(ZZ x EN) vs H(ZZ) x L(EN) & (11344, p < 0.001) \\
 L(ZZ x EN) vs H(ZZ) x L(EN) & (7096, p < 0.01)\\
\bottomrule
\end{tabular}%
\caption{\textbf{Experiment 4.} Results from a Mann-Whitney U-Test investigating whether the overlap percentages for different components across 7 models is significantly different for Jabberwocky and English hierarchical and linear grammars. We compare distributions of the mean overlap percentages for the top 1\% of MLP and attention components (p < 0.05, N = 108)}
\label{tab:expt4-stat-sig-jab}
\end{table*}

\end{document}